\documentclass{article}

\usepackage{microtype}
\usepackage{graphicx}
\usepackage{subcaption}
\usepackage{booktabs}

\usepackage{hyperref}

\usepackage[final]{colm2026_conference}

\usepackage{amsmath}
\usepackage{amssymb}
\usepackage{mathtools}
\usepackage{amsthm}
\usepackage[flushleft]{threeparttable}

\usepackage{longtable}
\usepackage{multirow}
\usepackage{array}
\usepackage{wrapfig}

\usepackage[capitalize,noabbrev]{cleveref}

\theoremstyle{plain}

\theoremstyle{definition}

\theoremstyle{remark}

\usepackage{xcolor}

\usepackage{comment}
\newcommand{\nummodels}{65}
\newcommand{\numproviders}{26}
\newcommand{\numtasks}{163}
\newcommand{\numbenchmarks}{27}
\newcommand{\numitems}{109,564}

\newcommand{\dataset}{WILD}
\newcommand{\model}{m}
\newcommand{\itm}{x}
\newcommand{\resp}{r}
\newcommand{\rec}{\langle \model, \itm, \resp \rangle}
\newcommand{\taskset}{\mathcal{T}}

\newcommand{\records}{\mathcal{R}}

\newcommand{\meanr}[1]{\mu_r(#1)}

\usepackage{pgfplots}
\pgfplotsset{compat=1.18}
\usepgfplotslibrary{units,groupplots}

\definecolor{irtblue}{HTML}{5B8DBE}       %
\definecolor{irtlightblue}{HTML}{8FB4D0}  %
\definecolor{mirtred}{HTML}{C85A54}       %
\definecolor{mirtlightred}{HTML}{DBA49E}  %
\definecolor{lrpurple}{HTML}{8B6DB3}      %
\definecolor{irt1plblue}{HTML}{A8D5F7}    %

\definecolor{uniformgray}{HTML}{BDC3C7}
\definecolor{anchorirtpurple}{HTML}{8E44AD}
\definecolor{anchorbinlavender}{HTML}{C5B3D9}
\definecolor{mtmorange}{HTML}{F0A860}     %
\definecolor{mtmeborange}{HTML}{E67E22}   %

\usepackage{lineno}

\definecolor{darkblue}{rgb}{0, 0, 0.5}
\hypersetup{colorlinks=true, citecolor=darkblue, linkcolor=darkblue, urlcolor=darkblue}

\title{Cost-Efficient Estimation of General Abilities Across \\Benchmarks}

\author{Michael Krumdick$^{1}$\thanks{Equal Contribution.} \;
Adam Wiemerslage$^{1}$\footnotemark[1] \;
Seth Ebner$^1$\;
Charles Lovering$^1$\\
{\bf  Chris Tanner}$^{1,2}$
\\
$^1$Kensho Technologies \quad $^2$MIT \\
\texttt{\{michael.krumdick,chris.tanner\}@kensho.com} \\
}

\begin{document}

\ifcolmsubmission
\linenumbers
\fi

\maketitle

\begin{abstract}

Thousands of diverse benchmarks have been developed to measure the quality of large language models (LLMs). Yet prior work has demonstrated that LLM performance is often sufficiently explained by a small set of latent factors, or \textit{abilities.} 
This suggests the potential for more efficient and principled benchmarking, but it remains difficult to compare the quality of different methods.
Motivated by predictive validity, we argue that the quality of a benchmarking framework should be grounded in how efficiently it enables the prediction of model performance on unseen tasks. 
To analyze this objective, we collect the ``Wide-scale Item Level Dataset" (\dataset), a dataset of item-model response pairs, comprising evaluations of \nummodels{} models on 109,564 unique items spanning {\numtasks} tasks drawn from {\numbenchmarks} datasets. This dataset enables the first analysis of how different techniques can predict a model's performance on a large, diverse collection of unseen tasks under different budget constraints. We demonstrate that combining a modified multidimensional item response theory (IRT) model with adaptive item selection driven by optimal experimental design can predict performance on 112 held-out benchmark tasks with a mean absolute error (MAE) of less than 7\% after observing only 16 items. We further demonstrate that incorporating cost-aware discount factors into our selection criteria can reduce the total tokens needed to reach 7\% MAE from 141,000 tokens to only 22,000, an 85\% reduction in evaluation cost.

\end{abstract}

\section{Introduction}
\label{sec:introduction}

A surge of research in LLM evaluation has resulted in thousands of benchmarks varied in domain and format, from grade-school word problems \citep{cobbe2021trainingverifierssolvemath} to completing code snippets \citep{zhuo2024bigcodebench} to answering multiple choice questions on foreign policy \citep{hendrycks2021measuring}. 
Despite the diversity, previous work has empirically demonstrated that the underlying ``capability space'' of language models is low-rank \citep{ruan2024observational,burnell2023revealingstructurelanguagemodel,Ili__2024}, meaning that many of these evaluations are measuring redundant capabilities. This manifests in high correlations between tasks \citep{ye-etal-2023-predictable}, creating the opportunity for developing ``compressed'' or efficient benchmarking techniques that leverage these relationships to reduce the total number of samples required to estimate performance \citep{perlitz-etal-2024-efficient,kipnis2025metabenchsparsebenchmark,pmlr-v235-maia-polo24a}.

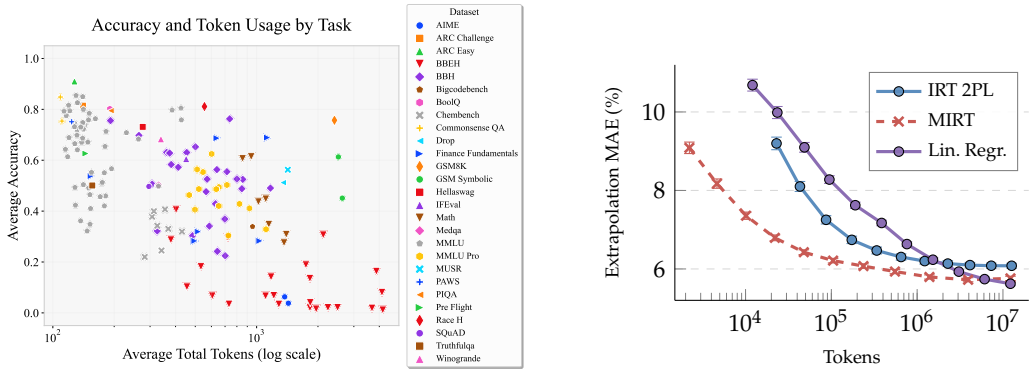
\begin{figure*}[t]
  \centering
  \begin{subfigure}[t]{0.5\textwidth}
    \centering
    \includegraphics[width=\textwidth]{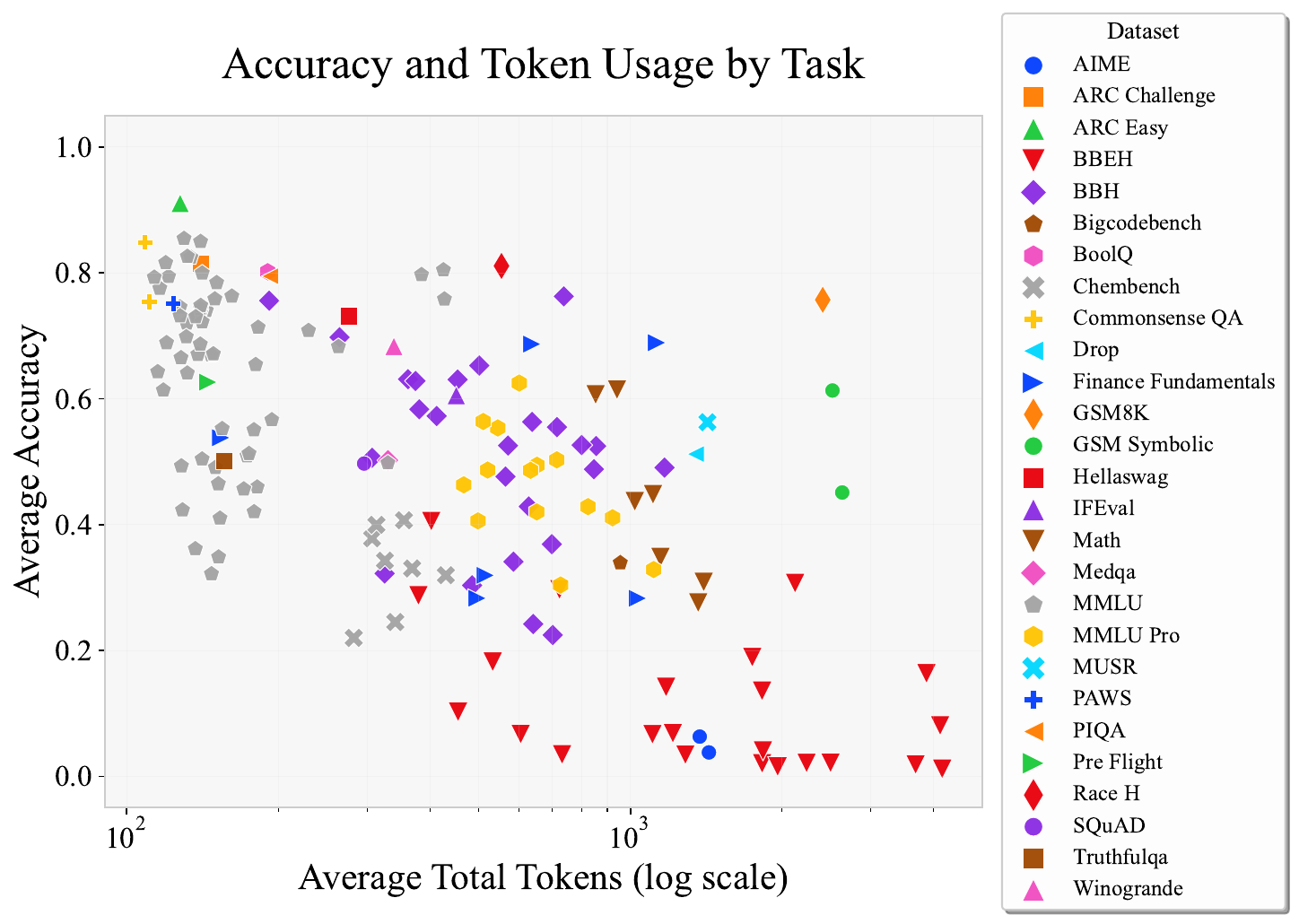}
  \end{subfigure}
  \hfill
  \begin{subfigure}[t]{0.45\textwidth}
    \centering
    \begin{tikzpicture}
    \begin{semilogxaxis}[
      width=\linewidth,
      height=0.75\linewidth,
      xlabel={Tokens},
      ylabel={Extrapolation MAE (\%)},
      xmin=1500, xmax=20000000,
      ymin=5.2, ymax=11.2,
      grid=major,
      grid style={dashed, gray!30, line width=0.4pt},
      ymajorgrids=true,
      xmajorgrids=false,
      legend style={
        at={(0.97,0.97)},
        anchor=north east,
        font=\scriptsize,
        draw=black!50,
        fill=white,
        fill opacity=0.9,
        text opacity=1,
        row sep=1pt,
        inner sep=2pt,
      },
      legend cell align={left},
      tick label style={font=\footnotesize},
      label style={font=\scriptsize},
      axis line style={black},
      every outer x axis line/.append style={black},
      every outer y axis line/.append style={black},
      separate axis lines,
      every outer x axis line/.append style={-},
      every outer y axis line/.append style={-},
      axis x line*=bottom,
      axis y line*=left,
    ]

    \addplot[
      color=irtblue, solid, line width=1.2pt,
      mark=*, mark size=1.8pt,
      mark options={fill=irtblue, draw=black, line width=0.4pt},
      error bars/.cd, y dir=both, y explicit,
        error bar style={solid, line width=0.4pt, color=irtblue},
        error mark options={rotate=90, mark size=2pt, line width=0.4pt, color=irtblue},
    ] coordinates {
      (22952.5,9.1982)+-(.0,0.1605)
      (43058.8,8.1012)+-(.0,0.1271)
      (86810.6,7.2518)+-(.0,0.1058)
      (173078.6,6.7411)+-(.0,0.0946)
      (337861.1,6.4700)+-(.0,0.0877)
      (649909.9,6.3043)+-(.0,0.0809)
      (1224624.2,6.2003)+-(.0,0.0758)
      (2276830.1,6.1364)+-(.0,0.0723)
      (4128087.2,6.0976)+-(.0,0.0681)
      (7313266.6,6.0818)+-(.0,0.0646)
      (12636134.2,6.0809)+-(.0,0.0624)
    };
    \addlegendentry{IRT 2PL}

    \addplot[
      color=mirtred, dashed, line width=1.2pt,
      mark=x, mark size=2.5pt,
      mark options={solid, line width=1pt, color=mirtred},
      error bars/.cd, y dir=both, y explicit,
        error bar style={solid, line width=0.4pt, color=mirtred},
        error mark options={rotate=90, mark size=2pt, line width=0.4pt, color=mirtred},
    ] coordinates {
      (2214.9,9.0860)+-(.0,0.1464)
      (4645.1,8.1741)+-(.0,0.1187)
      (10144.8,7.3562)+-(.0,0.0924)
      (21975.6,6.7925)+-(.0,0.0732)
      (47781.4,6.4267)+-(.0,0.0609)
      (104782.9,6.2120)+-(.0,0.0581)
      (236768.0,6.0725)+-(.0,0.0575)
      (551249.2,5.9337)+-(.0,0.0550)
      (1390407.2,5.7948)+-(.0,0.0534)
      (3927441.6,5.7249)+-(.0,0.0524)
      (12231620.0,5.7596)+-(.0,0.0547)
    };
    \addlegendentry{MIRT}

    \addplot[
      color=lrpurple, solid, line width=1.2pt,
      mark=*, mark size=1.8pt,
      mark options={fill=lrpurple, draw=black, line width=0.4pt},
      error bars/.cd, y dir=both, y explicit,
        error bar style={solid, line width=0.4pt, color=lrpurple},
        error mark options={rotate=90, mark size=2pt, line width=0.4pt, color=lrpurple},
    ] coordinates {
      (12063.6,10.6833)+-(.0,0.1510)
      (23466.6,9.9854)+-(.0,0.1535)
      (48552.4,9.0984)+-(.0,0.1239)
      (94908.4,8.2771)+-(.0,0.0985)
      (190048.0,7.6234)+-(.0,0.0791)
      (385412.7,7.1713)+-(.0,0.0696)
      (759592.3,6.6332)+-(.0,0.0586)
      (1527203.6,6.2350)+-(.0,0.0538)
      (3061408.3,5.9315)+-(.0,0.0491)
      (6113299.7,5.7404)+-(.0,0.0494)
      (12229504.6,5.6231)+-(.0,0.0495)
    };
    \addlegendentry{Lin.\ Regr.}

    \end{semilogxaxis}
\end{tikzpicture}
  \end{subfigure}
  \caption{\textbf{Left}: Token usage and accuracy averaged across all LLMs for the tasks in WILD. Tasks vary in token usage, leading to radically different item costs.
  \textbf{Right}: MAE for predicting the sample mean over 112 held-out tasks. The MIRT model with Optimal Experimental Design selection achieves low MAE for small sample size and token cost.}
  \label{fig:overview}
\end{figure*}

\textit{Task-oriented} evaluations attempt to estimate how well a model performs on a specific task or class of tasks. For example, one may want to estimate the probability that a model generates a correct pull request on real-world GitHub issues \citep{jimenez2024swebench}. In contrast, an \textit{ability-oriented} view of evaluation focuses on understanding how LLMs will perform across a broad range of (potentially nascent) tasks. For example, when benchmarking an LLM on MMLU \citep{hendrycks2021measuring}, the goal is often not to assess performance on multiple-choice questions. Instead, two key assumptions are made: (i) Answering these questions requires some set of task-independent abilities, and (ii) Performance on these questions measures those abilities \textit{and gives us information about how the LLM might perform on other tasks that require them} \citep{hernandez2017measure}.

While the quality of task-oriented evaluations can often be assessed by measuring how well they correlate with the desired behavior (e.g. how well does SWEBench predict a model's ability to close pull requests in practice), ability-oriented evaluations do not have a clear success metric. How, then, can one determine whether one set of evaluations is better than another? Psychometrics provides the framework of \textit{predictive validity}: the quality of an ability-oriented evaluation is determined by how well it predicts performance on other related tasks \citep{schellaert2025evaluation}. 

In this work, we develop a framework centered on predictive validity that evaluates the quality of model ability estimates. Prior work that uses predictive validity typically considers hand-curated task pairs. In contrast, we focus on efficiently predicting model performance across more than 100 held-out tasks. To enable this scale, we build the Wide-scale Item Level Dataset (\dataset{}): a large-scale dataset comprising 109,564 items from \numtasks{} distinct tasks derived from \numbenchmarks{} benchmarks, with responses from \nummodels{} LLMs for every item.

Dataset compression techniques often measure cost in terms of the number of items in the compressed dataset. While this may be reasonable for items from a single dataset, items across datasets can have radically different token-costs (\Cref{fig:overview}). The average item from the ``buggy tables'' scenario in Big Bench Extra Hard \citep{kazemi-etal-2025-big} compared to a ``world religions'' question from MMLU is 60 times more expensive to evaluate with \texttt{gpt-4o}. This observation motivates our primary focus: which methods can best predict performance across a large number of held-out tasks \textit{for the lowest total token cost}?

Our main contributions are:
\begin{enumerate}
\item We release \dataset{}, a wide-scale item-level dataset of model responses consisting of \nummodels{} models evaluated on \numitems{} unique items from \numtasks{} tasks.
\item We formulate a benchmark method comparison framework based on predictive validity, evaluating assessors on predicting model performance on over 100 benchmarks at a time.
\item We demonstrate that multidimensional IRT achieves $<$7\% MAE on held-out tasks after observing only 16 items, and $<6\%$ after observing 128 items, outperforming unidimensional IRT and simple baselines.
\item We introduce a cost-aware item-selection criterion that reduces evaluation token cost by over 80\% while maintaining prediction quality.
\end{enumerate}

\section{Related Work}

\noindent\textbf{Psychometrics in NLP} There is a burgeoning field of applying techniques from psychometrics to improve the evaluation of LLMs. \citet{rodriguez-etal-2021-evaluation} showed that IRT models rank LLMs more reliably than aggregate accuracy does, and that they can be used to select informative evaluation items. IRT models have been used more broadly to quantify the difficulty, discrimination, or other aspects of items and datasets \citep{yao2025the,vania-etal-2021-comparing,rodriguez-etal-2022-clustering}. Many works have tried to analyze the underlying structure of model performance using PCA \citep{ruan2024observational}, matrix factorization \citep{burnell2023revealingstructurelanguagemodel,Ili__2024}, performance predictability \citep{ye-etal-2023-predictable}, and human-informed rubrics of capability \citep{hendrycks2025definitionagi,zhou2025generalscalesunlockai}. Each of these works suggests that model performance is largely explained by a small number of factors. Our method combines a multidimensional IRT model with optimal experimental design guided item selection. Although these ideas have been explored for human testing \citep{mulder2009multidimensional,multidimadpative}, they have yet to be applied to LLM evaluation.

\noindent\textbf{Efficient Evaluation} 
Understanding an LLM's underlying abilities can yield practical benefits, such as reducing the cost of evaluations using metrics to evaluate subset quality \citep{perlitz-etal-2024-efficient}, or using clustering to select informative items \citep{vivek-etal-2024-anchor}. Towards this goal, many works explicitly use techniques from psychometrics.
\citet{pmlr-v235-maia-polo24a} propose tinyBenchmarks, which (1) adapts the clustering-based selection method anchor points to binary response patterns, and (2) develops IRT models to compress benchmarks, which are then used to estimate LLM abilities. \citet{kipnis2025metabenchsparsebenchmark} fit IRT models on HELM \citep{liang2022holistic} and use Fisher information-guided selection to create compressed versions of the benchmark. \citet{hofmann2025fluid} and \citet{li2025adaptive} use IRT models to create evaluation frameworks that use Fisher information to adaptively select the most informative items for a particular model, based on the model's previous responses. In contrast, other work has highlighted some of the limitations of psychometric techniques for efficient evaluation of LLMs \citep{Madaan2024QuantifyingVI}. 

\noindent\textbf{Ability-Oriented Benchmarking} Prior works have unified measures from different evaluations into estimates of general abilities. In computer vision, \citet{prabhu2024efficient} introduce a method for creating an ever-growing battery of both items and models. ONEBench \citep{ghosh2025onebench} builds an embedding-based retrieval system to aggregate items across benchmarks that target particular capability keywords. \citet{zhuang2024embedllm} introduce EmbedLLM, a method for learning descriptive embeddings for models inspired by matrix factorization. Most similar to our work is that of \citet{liao2025toward}, who utilize Bayesian MAP estimation and a structural prior to estimate latent abilities across benchmarks. However, these approaches either focus on task-specific compression or do not evaluate cross-task predictive validity with more than a handful of tasks. 

\section{Preliminaries}

In this section, we provide a brief overview of the main IRT models used in this work, along with the optimal experimental design criteria used for selection.

\subsection{Item Response Theory}

A \textit{record} is a tuple consisting of a model, item, and response: $(\model, \itm, \resp)$. In our work, every response $r$ is represented as a binary label, $0$ or $1$, that indicates the correctness of the model's prediction. 
Given a dataset of records, $\records = \{(\model_0, \itm_0, \resp_0), \dots \} $, we would like to learn an \textit{assessor} that can predict if an LLM $m$ will produce a correct response for a given item $x$. The assessor's prediction is defined as $p(r = 1 | \model, \itm)$. Psychometrics provides a family of models, originally developed for human testing, suited for this purpose.
Specifically, IRT uses both the test-taker's ability $\theta_{\model} \in \mathbb{R} $ and the item's difficulty $\delta_{\itm} \in \mathbb{R}$ to model the ability of a test-taker to correctly answer an item:
\begin{equation*}
    p(r = 1 | \model, \itm) = \sigma(\alpha_{\itm} ( \theta_{\model} - \delta_{\itm} )) = \frac{1}{1 + e^{-\alpha_{\itm} ( \theta_{\model} - \delta_{\itm} )}}.
\end{equation*}
The above model, referred to as a two-parameter logistic (2PL) model, has scalar parameters that capture both an item's difficulty $\delta_\itm$ and its discrimination $\alpha_{\itm}$. A one-parameter logistic (1PL) model fixes $\alpha_{\itm} = 1$.

We can extend the above framework to a Multidimensional IRT Model (MIRT): rather than defining ability as a single scalar value $\theta_{\model}$, one can represent model $m$'s abilities as a vector $\Theta_\model \in \mathbb{R}^{d}$, where $d$ is the number of abilities. Each item $x$  will have a \textit{loading} vector $K_{\itm} \in \mathbb{R}^{d}$ that weights the individual $\Theta_\model$ dimensions. The model's ability for item $x$ is defined as: %
\begin{equation}
\label{eq:mirt_logistic}
    p(r = 1 | \model, \itm) = \sigma(\alpha_\itm (\langle \Theta_{\model}, K_{\itm} \rangle - \delta_{\itm} )).
\end{equation}

\subsection{Item Selection}
\label{eq:item_selection}

IRT models can \textit{adaptively} select items for assessment based on Fisher information, $\mathcal{I}$. The Cramer-Rao bound states that for an unbiased estimator, the variance is at least $1/\mathcal{I}(\theta_m)$:
\begin{equation}
\label{eq:cramer_rao}
\text{Var}(\hat{\theta}_m) \ge \frac{1}{\mathcal{I}(\theta_\model)}.
\end{equation}
Fisher information is additive over independent samples. For the 2PL model:
\begin{equation}
    \label{eq:2pl_fisher_info}
  \mathcal{I}_{\model,\itm} = \alpha_\itm^2 \sigma(\alpha_{\itm} ( \theta_{\model} - \delta_{\itm} )) (1 - \sigma(\alpha_{\itm} ( \theta_{\model} - \delta_{\itm} ))).
\end{equation}
This quantity is maximized when $p(r=1|\model,\itm) = 0.5$. Adaptive item selection proceeds by choosing items with the highest Fisher information for the current ability estimate, which equivalently corresponds to the item with the highest expected reduction in variance.

Instead of a single Fisher information value, the multidimensional case requires analysis of a Fisher information \textit{matrix}.
Given a loading vector $K_{\itm}$, we modify \Cref{eq:2pl_fisher_info} to the multidimensional case to get:
\begin{equation}
\label{eq:2pl_mirt_fisher_info}
p = \sigma(\alpha_{\itm} ( \Theta_{\model}^{ T} K_{\itm} - \delta_{\itm} )), \qquad
\mathcal{I}_{\model, \itm} = \alpha_\itm^2  p(1 - p)K_{\itm} K_{\itm}^{T}.
\end{equation}

When $|K| > 1$, we have a Fisher information matrix. 
The diagonals of this matrix represent the marginal information an item carries for $\Theta_{m,i}$ if its response is observed. The off-diagonals represent the joint information between $\Theta_{m,i}$ and $\Theta_{m,j}$.
Selecting items with the maximal trace for this matrix is analogous to maximizing the Fisher information.
However, while maximizing the Fisher information is equivalent to minimizing the variance of $\hat{\theta}$ in the single-dimensional case, this is not true with a MIRT model, where the variance of $\hat{\Theta}$ depends on the full matrix structure.
To address this limitation, we borrow from the Optimal Experimental Design (OED) literature \citep{kiefer1959optimum} to develop selection objectives that model the variance of a multidimensional $\hat{\Theta}$.

\noindent\textbf{Optimal Experimental Design}
OED criteria typically model the inverse of the accumulated Fisher information after observing an item:
\begin{equation}
(\mathcal{I}_{\textrm{cumulative}} + \mathcal{I}_{\model, \itm})^{-1}.
\end{equation}
\noindent where $\mathcal{I}_\textrm{cumulative}$ is the current cumulative sum of the information contributed from each item. Given the Cramer-Rao bound in \Cref{eq:cramer_rao}, this acts as a lower bound on the estimator variance.
Many popular OED criteria focus on minimizing a function of the information matrix parameters.
We found that V-optimality, which minimizes the prediction variance over observed points, 
works best in preliminary experiments.
Concretely, we define the \textit{prediction matrix} $M = \sum_{x \in \mathcal{X}_t} \alpha_x^2 K_x K_x^\top$, where $\mathcal{X}_t$ is the set of items belonging to target benchmarks,  whose parameters are known from training but whose responses for $\model$ are to be predicted. Given the current posterior covariance  $C = \mathcal{I}_{\textrm{cumulative}}^{-1}$, the V-optimal selection score for candidate item $\itm$ is:
\begin{equation}
    \label{eq:voptimal}
    s_\itm = \frac{w_\itm \alpha_\itm^2 \, c_\itm^\top M c_\itm}
              {1 + w_\itm \alpha_\itm^2 \, K_\itm^\top c_\itm}, 
\quad c_\itm = C K_\itm, \quad w_\itm = p_\itm(1-p_\itm).
\end{equation}
After each observation, the posterior covariance is updated via the Woodbury correction (\Cref{sec:woodbury_correction}).
V-Optimality thus prioritizes items whose loadings align with $M$, which we expect to prioritize 
``general" factors that are informative for predictive ability across tasks.

\section{\dataset{} Dataset}
\captionsetup{skip=4pt}                    %
\setlength{\aboverulesep}{0pt}
\setlength{\belowrulesep}{0pt}
\setlength{\tabcolsep}{4pt}

\begin{wraptable}{R}{0.42\textwidth}   %
\vspace{-\intextsep}                   %
\centering
\scriptsize
\caption{Comparison of \dataset{} to other item-level response datasets.}
\label{tab:comp}
\begin{threeparttable}
\begin{tabular}{lrrr}
\toprule
 & Models & Tasks & Items \\
\midrule
\citet{perlitz-etal-2024-efficient}        & 44    & 16 & 65,000  \\
\citet{kipnis2025metabenchsparsebenchmark} & 5055\tnote{*} & 63 & 28,632 \\
\citet{zhou2025generalscalesunlockai}      & 15    & 63 & 16,108  \\
\citet{liao2025toward}                     & 70    &  5 &  8,918  \\
\midrule
\dataset{}  & \nummodels{} & \numtasks{} & \numitems{} \\
\bottomrule
\end{tabular}
\begin{tablenotes}[flushleft]
  \item[*] Includes model variants.
\end{tablenotes}
\end{threeparttable}
\end{wraptable}

To support the ability to estimate LLM performance across diverse tasks, we needed a large dataset of model results on many tasks.
Existing item-level response datasets contain limited individual tasks or have other quality issues. For example, the OpenLLM Leaderboard \citep{open-llm-leaderboard-v2} contains thousands of different models, but most are adapted from a small set of base models. HELM \citep{liang2022holistic} contains many tasks but lacks item coverage for many models.

\dataset{} represents the largest complete cross-task evaluation dataset where all model-task pairs were evaluated using a single consistent framework. \Cref{tab:comp} compares this dataset to other item-level response datasets.\footnote{\dataset{} is available at \url{https://huggingface.co/datasets/kensho/WILD}}

We collect \numitems{} items across \numtasks{} tasks from \numbenchmarks{} benchmarks. We focused on high quality tasks with compatible implementations \citep{abbas2025developing}, including:
multiple choice \citep[MMLU Pro]{wang2024mmlu}, math \citep[MATH]{hendrycksmath2021}, coding \citep[bigcodebench]{zhuo2024bigcodebench}, and domain-specific \citep[BizBench]{krumdick-etal-2024-bizbench} evaluations.

\dataset{} includes models from 0.5B parameters to large, closed API models. For our main analysis, we exclude explicit reasoning models, motivated by the increased costs and additional hyperparameter complexities introduced by this family of models (reasoning effort, number of reasoning tokens, etc.). In   
\Cref{ref:reasoning_study}, we provide a preliminary analysis that suggests our results apply to reasoning models; however, we leave a 
larger investigation for future work. We evaluate all models using the Inspect AI framework \citep{UK_AI_Security_Institute_Inspect_AI_Framework_2024}. In total, we evaluated \nummodels{} unique models, sourced from \numproviders{} different organizations. A full list of tasks and models included in \dataset{} can be found in 
\Cref{sec:detailed-model-task}.

\section{Methods}
\label{sec:methods}

Our primary method for cross-task ability estimation modifies a bifactor model \citep{holzinger1937bi}.
We describe the model in \Cref{sec:methods:nested} and the fitting procedure in \Cref{sec:methods:mirt}. Finally, in \Cref{sec:methods:cost}, we incorporate token costs into the selection methods.

\subsection{Nested MIRT Model}\label{sec:methods:nested}

We modify the standard bifactor model---where each item loads on a general factor and exactly one task specific factor---by replacing the task specific factors with latent factors, akin to the MIRT model,
\begin{equation}
\label{eq:mirt_logistic_g}
    p(r = 1 | \model, \itm) = \sigma(\alpha_\itm (\Theta_{\model, g} + \langle \Theta_{\model}, K_{\itm} \rangle - \delta_{\itm} )).
\end{equation}
This ensures that a single factor remains active for all items, thereby creating a shared general factor.
The utility of such a general factor in ability estimation is well-established in the literature \citep{spearman1904general}, and we confirm this in a preliminary factor analysis in \cref{fig:combined} in the appendix.
To emphasize the importance of the general factor, we optimize a regularized log-likelihood that encourages shrinkage toward a nested unidimensional model, improving stability in low-sample regimes. Let $\ell(r, z)$ denote the Bernoulli log-likelihood with logit $z$,
\begin{equation}
\ell(r, z) = r \log \sigma(z) + (1 - r)\log(1 - \sigma(z)).
\end{equation}
We extract two logits from the model and combine both for our loss:
\[
z_{\text{full}} = \alpha_\itm (\Theta_{\model, g} + \langle \Theta_{\model}, K_{\itm} \rangle - \delta_{\itm}), \quad
z_g = \alpha_\itm (\Theta_{\model, g} - \delta_{\itm}), \quad
\mathcal{L}_{\text{total}} = \ell(r, z_{\text{full}}) + \lambda \, \ell(r, z_g).
\]

We include an ablation and analysis of these modifications in \Cref{sec:MIRT ablations}, and refer to this implementation as MIRT in the remainder of the paper.

\subsection{Fitting the MIRT Model}\label{sec:methods:mirt}

Our fitting procedure uses coordinate gradient descent with L-BFGS-B \citep{byrd1995limited} on the model and item parameters. We include a standard normal prior on $\Theta$. We use this fitted MIRT model to efficiently form estimates of held-out LLMs' $\Theta$ via maximum a posteriori (MAP) estimation on the frozen item parameters from the available records. In order to learn stable updates from very small numbers of items, we include an empirical prior in the MAP update, $\Theta_\model \sim \mathcal{N}(\mu_{\Theta}, \Sigma)$, where $\mu_{\Theta}$ is the mean of the observed $\Theta$ parameters and $\Sigma$ is the covariance matrix between the dimensions of $\Theta$. We then maximize
\begin{align}
\log p(\Theta_\model | \itm, r) &\propto \mathcal{L}_{\text{prior}}(\Theta_\model) + \log p(r | \model, \itm).
\label{eq:posterior}
\end{align}

More details on the training and MAP process can be found in \Cref{sec:training-details}.

\subsection{Selecting Items with Token-Aware Discounting}\label{sec:methods:cost}

Prior work explores efficient evaluation with the goal of reducing the number of samples an LLM must respond to, effectively considering the cost of all items to be equal. However, items often vary greatly in cost. \Cref{fig:overview} demonstrates the variation in total tokens across tasks in WILD. 
This translates directly into compute and API cost. We thus incorporate a cost term 
to encourage selection of cheap items:
\begin{equation}
  \overline{\tau}_{\itm} = \frac{1}{|\mathcal{M}_{train}|}\sum_{\model \in \mathcal{M}_{train}} \tau(\model, \itm),
\end{equation}
\noindent where $\mathcal{M}_{train}$ is the set of all models used for training, and $\tau(\cdot)$ counts the sum of input tokens encoded by and output tokens generated by $\model$ for $\itm$.
We can then modify any information based selector 
to get a ratio between (a) the information gain from observing an item, and (b) the item's expected cost for an unknown model: 
$\mathcal{I}_\itm / \overline{\tau}_{\itm}$.
Intuitively, this identifies items with 
the best tradeoff between informativeness and cost.
We denote cost-aware selectors with the -$\tau$ suffix.

\section{Experimental Setup}\label{ref:experimental_setup}

We formulate our framework as item-level prediction over a dataset with a large number of records across a series of models and tasks. 
The item-level prediction models, or \textit{assessors}, aim to infer how a model will perform on a set of held-out items given its observed performance on a separate set of items. 
We consider two types of prediction:
\textbf{Interpolation} predicts performance on held-out items from the \textit{observable tasks}. \textbf{Extrapolation} predicts performance on held-out items from a \textit{different set of tasks}.

\subsection{Splits}
We split the data at two levels. First, models are partitioned into train and test sets. 
For the train models, the assessor is allowed to observe performance on all items across all tasks.
In our experiments, we perform cross-validation over 20 random model partitions, each consisting of 55 train models and 10 test models.
We additionally randomly partition tasks into interpolation and extrapolation sets. The assessor observes the performance of each test model on items from the interpolation tasks. The assessor then predicts performance on held-out items from those same tasks (interpolation), as well as performance on items from unseen tasks (extrapolation).
For each task, we hold out 128 items and filter out tasks with fewer than 128 items for computing prediction metrics.

More formally, the \textbf{calibration} records consist of a set of test models $\mathcal{M}_{test}$ for evaluation (all of which are previously unobserved) with a subset of historic tasks, $\mathcal{T}^c \subset \mathcal{T}$, containing items $\mathcal{X}^c$.
The \textbf{validation} records $\mathcal{X}^v$ are also defined over $\mathcal{M}_{test}$, but they comprise a set of extrapolation tasks $\mathcal{T}^v \subset \mathcal{T}$ that are disjoint from the interpolation tasks: $\mathcal{T}^v \cap \mathcal{T}^c = \emptyset$.
See \cref{tab:data_partitioning} in the Appendix for an overview of these splits.

\begin{figure}[t]
  \centering
\includegraphics[width=\linewidth]{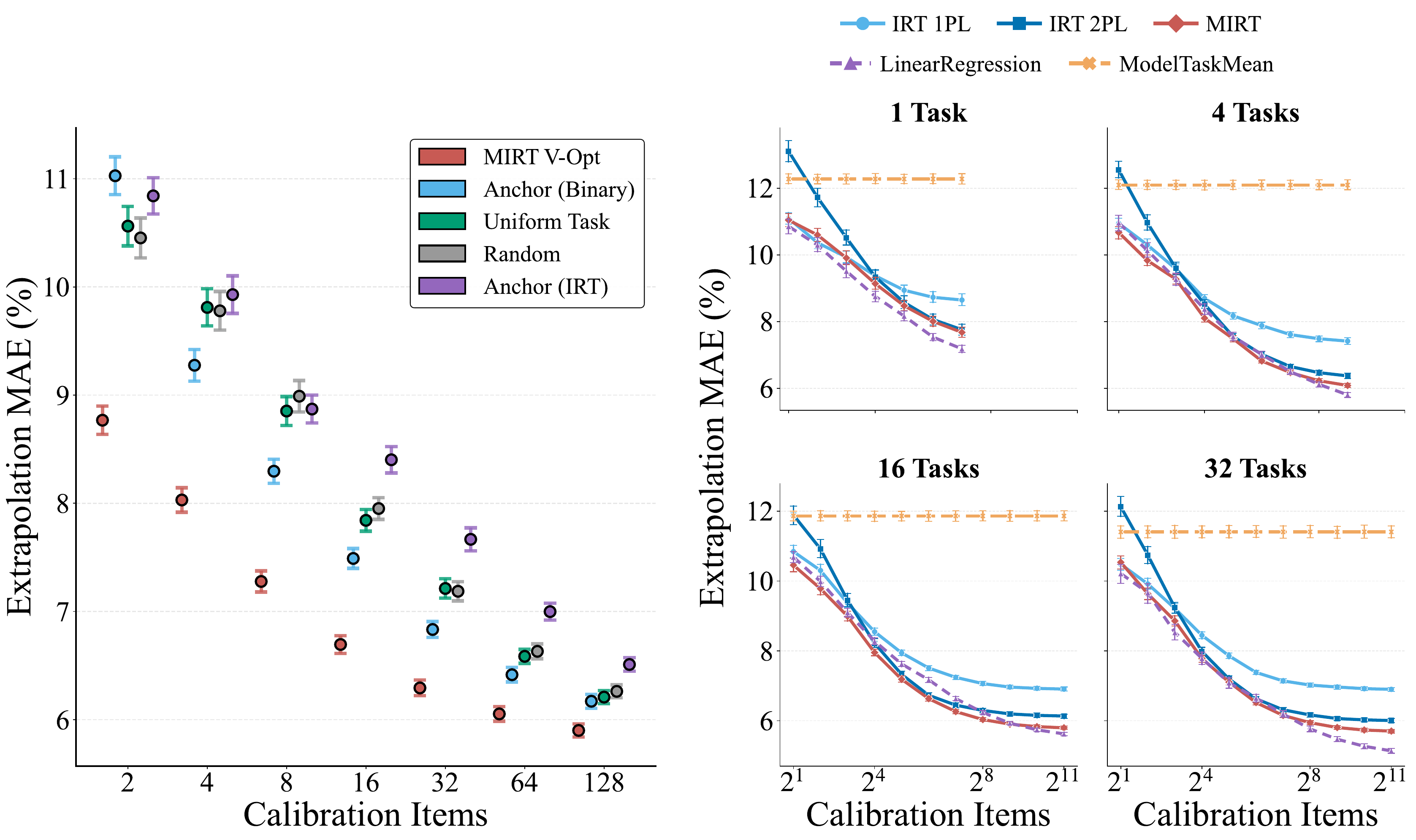}
  \caption{ \textbf{Left}: Baseline selectors vs. Adaptive V-Optimal. \textbf{Right}: Prediction comparison. MIRT v.s. baseline predictors as we increase the number of extrapolation tasks. Model Task Mean (MTM) baselines are constant because extrapolation tasks are fully held out for the test models. See \Cref{fig:calib_valid_full} for interpolation results.}
  \label{fig:mirt_baselines_and_prediction}
\end{figure}

\subsection{Evaluation Paradigms}
\label{sec:eval_paradigms}
We evaluate how well the assessor predicts an unseen model's performance when provided only $N$ samples from the calibration records. We perform three types of evaluation based on how the sampled calibration records are provided to the assessor.

\textbf{Prediction} The assessor receives a random sample of $N$ items from the calibration records.
This evaluates how well the assessor predicts model performance from a static benchmark.

\textbf{Selection} The assessor selects the $N$ items from the calibration records all at once. This evaluates how well the assessor can compose static benchmarks out of a pool of items.

\textbf{Adaptation} The assessor iteratively selects $N$ items for each calibration model, conditioned on how the model has performed on previously selected items. This evaluates how well the assessor can create adaptive benchmarks for a given model. 

For each task, we hold out a set of items for computing the evaluation metrics. Model performance on these held-out items can never be observed by the assessor.
This is a subtle distinction with prior work which evaluates dataset compression performance by how well the model predicts the dataset performance mean from items that were used to compute that mean. Importantly, assessors within such a framework will always reach perfect performance once they observe all items. In contrast, our evaluation involves some amount of irreducible error. We provide further discussion in \cref{sec:held-out}.
This makes exactly one interpolation setting in our experiments comparable to prior work on benchmark compression. We present that setting in isolation in \Cref{tab:interpolation_mae} in the appendix.

\subsection{Metrics}

Although we require assessors to predict item-level performance, our main performance results evaluate aggregate accuracy over tasks. 
Item-level metrics may better capture how well the assessor can predict how a model will perform, but task-level metrics are generally more interpretable and align with how most practitioners interact with benchmarks.

\textbf{Task-level MAE} measures how well the predictor captures task-level performance:
\begin{equation}
  \text{MAE}_\text{task}(a, \records) = \frac{1}{|\taskset|}\sum_{t \in \taskset} \left|\meanr{\records_t} - \mu_a (\records_t) \right|,
\end{equation}
averaged across all tasks $\taskset$, where $\meanr{\records_t}$ is the empirical mean over the records and $\mu_a (\records_t)$ is the mean of the assessor prediction of $p(r = 1)$ over the records.

\textbf{Token Cost} measures the cost of the $N$ samples, computed as the sum of the input and output tokens across all items.
In practice, because encoding costs are cheaper than generation costs in LLMs, 
we halve the cost of input tokens.

\subsection{Prediction Baselines}
\label{sec:exp_baselines}
\noindent\textbf{Sample Mean} This is the simplest baseline predictor. From a set of records, we estimate:
\begin{equation}
\hat{\theta}_{\model, t} = \meanr{\records_t} = \frac{1}{|\records_t|} \sum_{\rec \in \records_t} \mathbb{I}(\resp = 1).
\end{equation}
We implement a simple predictor, $p_{\text{mean}}(r = 1 | \model, \itm_t) = \hat{\theta}_{\model, t}$, which predicts performance for an unseen item based on the model sample mean for that task. 
However, when $|\records_t|$ is small, sample mean estimation may be poor.
We develop a Bayesian extension using the training data as a prior and use Bayesian shrinkage on our estimate.
For more details on the Bayesian estimate and results for sample mean baselines, see \Cref{sec:interpolation_results}.

\noindent\textbf{Linear Regression}
We fit a ridge regression model for each task from the vector of responses for $\mathcal{M}_{train}$ on observed items to the overall mean performance on that task.
Then we apply each task-specific model to $\mathcal{M}_{test}$.
We train a new model for every set of calibration item observations.
This is a strong and somewhat privileged baseline: it explicitly trains a new model to predict each target task, whereas IRT models learn a task-agnostic latent ability space with no knowledge of which tasks will be evaluated at test time, making the large set of evaluation tasks harder for the IRT models.

\noindent\textbf{2PL IRT} We include a single dimensional, standard IRT model trained across all items. We fit this model using the same method as the MIRT model. With Fisher selection, this matches the setup used by \citet{hofmann2025fluid} in the Adaptation setting. In the Selection setting, it matches the method from \citet{kipnis2025metabenchsparsebenchmark}. However, since our model universe is smaller than \citet{kipnis2025metabenchsparsebenchmark}, we use a single median $\theta$ rather than binning across all possible $\theta$.

\subsection{Selection Baselines}
\textit{Selectors} are responsible for choosing items for the \textbf{Selection} and \textbf{Adaptation} evaluations. We include some simple selection baselines. \textbf{Random} selects randomly from the pool of items. \textbf{Uniform} evenly allocates items over each task, randomly selecting tasks to receive the remainders in the likely case that the number of items is not divisible by the number of tasks. To test efficiency, \textbf{Min Cost} (See \Cref{fig:min_cost} for results) selects the items with the lowest expected cost based on the estimator described in \Cref{sec:methods:cost}.

\noindent\textbf{Anchor Points} We compare with the anchor points \citep{vivek-etal-2024-anchor} implementations described in \citet{pmlr-v235-maia-polo24a}, where items are clustered with $k$-means and points close to the centroids are selected.
We compare to two of their implementations that differ in the item representation:
\textbf{Anchor-binary} represents each item as the vector of all of its historic model responses.
\textbf{Anchor-IRT} instead represents items as the concatenation of their loading and difficulty parameters from the IRT predictor.

\begin{figure*}[t]
  \centering
\input{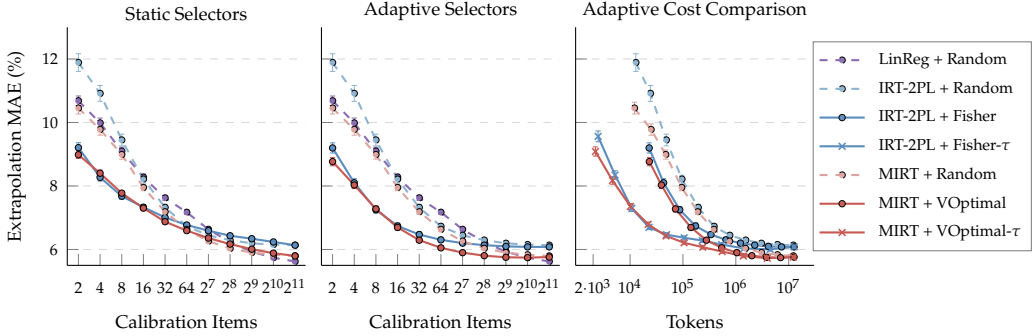}
  \caption{Summary of main results comparing the performance of the Nested MIRT and 2PL IRT model. (Left) Comparison in the static selection setting. (Center) Comparison in the adaptive selection setting. (Right) Comparison of token efficiency in the adaptive setting. In all three plots, Random selection corresponds to the Prediction setting (\Cref{sec:eval_paradigms}).}
  \label{fig:static_adaptive_cost}
\end{figure*}

\section{Results \& Analysis}
\label{sec:results}
For all results we compare to the Nested MIRT model with $K$=3. This choice of $K$ is based on a factor analysis presented in \Cref{sec:factor_analysis}. See \Cref{sec:MIRT ablations} for full post-hoc ablations of $K$.

\subsection{Prediction}

\Cref{fig:mirt_baselines_and_prediction} compares the extrapolation performance of MIRT to all baseline predictors on the right. We vary the number of calibration tasks as a proxy for item diversity. Model performance tends to increase as the number of tasks increases even as the samples per task decreases. 
This is especially consistent for the MIRT model: as the number of tasks and samples increases, the improvements of MIRT over the IRT-2PL predictor also increases. This provides an interesting scaling direction. For a fixed number of samples, selecting from a larger set of tasks increases the efficiency of estimating LLM abilities with MIRT.
The regression model performs best in the single task case, and  when there are large numbers of calibration items.
This highlights the IRT model's ability to generalize latent capability representations from other tasks.
It also demonstrates that at a high enough number of calibration samples, the regression assessor, which fits a new model to predict each individual task and calibration item set, is unsurprisingly extremely well-calibrated.
The IRT models may suffer from noise in the difficult cross-task generalization that they must solve.
However, we show in \Cref{sec:results_selection} that sampling high-information items under the IRT model solves this issue, leading to better performance than the regression model.

In \Cref{sec:interpolation_results} we present interpolation results, where the differences between IRT models are much smaller than in the extrapolation setting.
Interpolation is similar to prior work that focused on compressing a given benchmark, particularly at the 1-task setting, which highlights the difficulty of extrapolation, where we see more variance in the results.

\subsection{Selection}
\label{sec:results_selection}
In \Cref{fig:static_adaptive_cost} (left) we compare 
MIRT, IRT, and regression assessors with both random and information-based selectors.
At low numbers of calibration items, the Fisher selector for the IRT model reduces MAE significantly compared to random selection, and performs on-par with MIRT V-Optimal until 32 calibration items. 
The V-Optimal selector for MIRT achieves consistently strong performance compared to other methods until the regression baseline surpasses it at 1024 ($2^{10}$) samples.

\subsection{Adaptation}
\Cref{fig:static_adaptive_cost} (center) presents results for the adaptive setting, where items are selected one at a time per LLM, adapting to the current ability estimate. %
 At all but the maximum calibration item budgets, adaptive selection improves results for V-Optimal and Fisher selectors, with the adaptive MIRT V-Optimal assessor outperforming the IRT-2PL Fisher assessor by a larger gap (see \Cref{fig:adaptive_v_static} for a direct comparison).
In \Cref{fig:static_adaptive_cost} (right), we investigate the token efficiency of each method.
Here, the $x$-axis is the total number of tokens after sampling $N$ calibration items at the same intervals as the other two plots.
Token counts are averaged over all seeds and splits.
For both the MIRT and IRT models, token-aware discounting leads to significant savings, reducing MAE to $<$7\% with fewer than 50,000 tokens versus almost 400,000 tokens with random selection. 
\Cref{fig:mirt_baselines_and_prediction} (left) compares the performance of various baseline selectors for the MIRT model. Adaptive V-Optimal consistently outperforms the baselines. Once $N \approx 64$, each of the baseline selection methods performs roughly on par with random selection.

A natural question given the token-efficient results is: do we actually need to select by Fisher information if we can simply select the cheapest possible items?
In \Cref{sec:min_cost_selection}, we demonstrate that the cheapest items in terms of tokens do not lead to efficient results, and the Fisher information based selection is crucial.

\section{Conclusion}

We demonstrate that for large numbers of held-out tasks, IRT methods often greatly improve task-level prediction accuracy, in comparison to several strong baselines. Our approach is motivated by and leverages both psychometric literature \citep{lord1968statistical} and Optimal Experimental Design \citep{kiefer1959optimum}. We introduce assessor models that are effective and cost-efficient not only in the base prediction task---predicting LLM performance on held-out items---but also in selecting assessment items out of a pool of benchmarks. A key contribution of this work is a novel cost-aware term which leads to 
high quality model ability estimates with an order-of-magnitude reduction in token costs.

\section*{LLM Use Disclosure}
Claude Code was used extensively to write several parts of the experimental framework, and to help with code for generating plots.

\bibliography{references}
\bibliographystyle{colm2026_conference}

\newpage
\appendix
\crefalias{section}{appendix}
\crefalias{subsection}{appendix}

\section{Training Details}
\label{sec:training-details}

\paragraph{MIRT Fitting}
We maximize the log-likelihood of \Cref{eq:mirt_logistic}:
\begin{equation}
    \ell = \sum_{(\model, \itm, \resp) \in \records_{\textrm{train}}} \resp \log p_{\model\itm} + (1 - \resp)\log(1-p_{\model\itm})
\end{equation}
where $p_{\model\itm} = \sigma(\alpha_{\itm}(\Theta_{\model}^\top K_{\itm} - \delta_{\itm}))$.

We add a standard normal prior on abilities:
\begin{equation*}
    \log p(\Theta) = -\frac{1}{2} \sum_{\model,d} \Theta_{\model d}^2
\end{equation*}
and an $\ell_1$ penalty on loadings for sparsity. The objective is:
\begin{equation*}
    \mathcal{L} = \ell + \log p(\Theta) - \lambda \sum_{\itm,d} |K_{\itm d}|
\end{equation*}

Defining residuals $e_{\model\itm} = p_{\model\itm} - \resp$ for each observed record, and indexing parameter dimensions with d, we compute the gradients:
\begin{align*}
     \frac{\partial \mathcal{L}}{\partial \delta_{\itm}} &= -\sum_{\model} \alpha_{\itm} \cdot e_{\model\itm} \\
     \frac{\partial \mathcal{L}}{\partial \Theta_{\model d}} &= \sum_{\itm} \alpha_{\itm} \cdot e_{\model\itm} \cdot K_{\itm d} + \Theta_{\model d} \\
     \frac{\partial \mathcal{L}}{\partial K_{\itm d}} &= \sum_{\model} \alpha_{\itm} \cdot e_{\model\itm} \cdot \Theta_{\model d} + \lambda \, \text{sign}(K_{\itm d})\\
     \frac{\partial \mathcal{L}}{\partial \alpha_{\itm}} &= \sum_{\model} (\Theta_{\model}^\top K_{\itm} - \delta_{\itm}) e_{\model\itm}
\end{align*}

We optimize the parameters jointly using a coordinate descent algorithm.

\paragraph{MAP update}

We update $\hat{\Theta}_\model$ for new models by optimizing \Cref{eq:posterior} with coordinate descent.
We initialize new $\hat{\Theta}_\model$ as the mean $\Theta$ from $\mathcal{M}_{\textrm{train}}$ after fitting.

The gradient with respect to the ability parameters is given by:
\begin{align*}
\dfrac{\partial }{\partial \Theta_\model} \log p(\Theta_\model | \itm, r) =& - \Sigma^{-1} (\Theta_\model - \mu_{\Theta}) \\
& + \dfrac{\partial }{\partial \Theta_\model} \log p(r | \model, \itm)
\end{align*}

We compute this over all items selected by the assessor, while keeping their parameters frozen.

For both phases of optimization, we use the minimize function implemented in SciPy \citep{virtanen2020scipy}.
We clip probabilities to the range $[\epsilon, 1 - \epsilon]$
where $\epsilon=10^{-10}$ to avoid numerical instability in the logarithm.
For fitting on the full historic set of records, we use the L-BFGS-B implementation for up to 1000 steps.
For MAP estimation, we use the more precise trust-ncg algorithm for up to 100 steps.

\subsection{Degenerate Items}
Certain items in a training set may be \textit{degenerate}.
That is, they may be constant in terms of model responses, containing all correct (or all incorrect) responses for every single model in the training split.
This may happen in cases where an item is too hard or otherwise impossible due to errors.
This can also occur for an item that is too easy, perhaps having been leaked into LLM training data.
These items cause numerical stability issues for learning IRT models.
We thus filter out degenerate items at runtime during training.
At inference, when an IRT model is asked to make a prediction for a degenerate item, we then have no information for that item.
To solve this, all models will heuristically predict the majority outcome for that item in the training data (correct or incorrect).
This intuitively works quite well as these items will always have full agreement across train models.

\subsection{Woodbury Correction}
\label{sec:woodbury_correction}
When doing V-Optimal selection, we need to update the posterior covariance after each observation.
However, computing the inversion $\mathcal{I}_{\textrm{cumulative}}^{-1}$ is expensive.
Instead, we apply the Woodbury matrix identity to perform a rank-1 update directly to C:
\begin{equation}
    \label{eq:woodbury_update}
    C \leftarrow C - \frac{w_\itm \, \alpha_\itm^2 
    C K_\itm K_\itm^\top C}{1 + w_\itm \, \alpha_\itm^2 
    K_\itm^\top C K_\itm}
\end{equation}
This enables $O(d^2)$ sequential updates without recomputing $\mathcal{I}_{\textrm{cumulative}}^{-1}$ at each step.

\section{Additional Prediction Results}
\label{sec:interpolation_results}
\paragraph{Model Task Mean Implementation}
The quality of the sample mean baseline described in \Cref{sec:exp_baselines} depends on the number of samples that we have per task.
As $|\records_t| \rightarrow \infty $, this is the optimal predictor for the interpolation task.
However, when $|\records_t|$ is small, it performs rather poorly. 
The Bayesian extension introduced in \Cref{sec:exp_baselines} resolves this with Bayesian shrinkage on the estimate with a Beta-Binomial prior.
For each task, we estimate the mean performance over all train models $\mu_{\mathcal{M}, t}$:
\begin{equation}
\hat{\theta}_{\model, t} = \frac{\alpha + \sum_{\rec \in \records_t} \mathbb{I}(\resp = 1)}{\alpha + \beta + |\records_t| },
\end{equation}
where $\alpha = \lambda\, \mu_{\mathcal{M}, t}$ and $\beta = (1 - \mu_{\mathcal{M}, t})\, \lambda$ and $\lambda$ is a parameter used to control the shrinkage strength. We set $\lambda = 5$ based on validation experiments.

\begin{figure*}[t]
  \centering
  \includegraphics[width=0.9\textwidth]{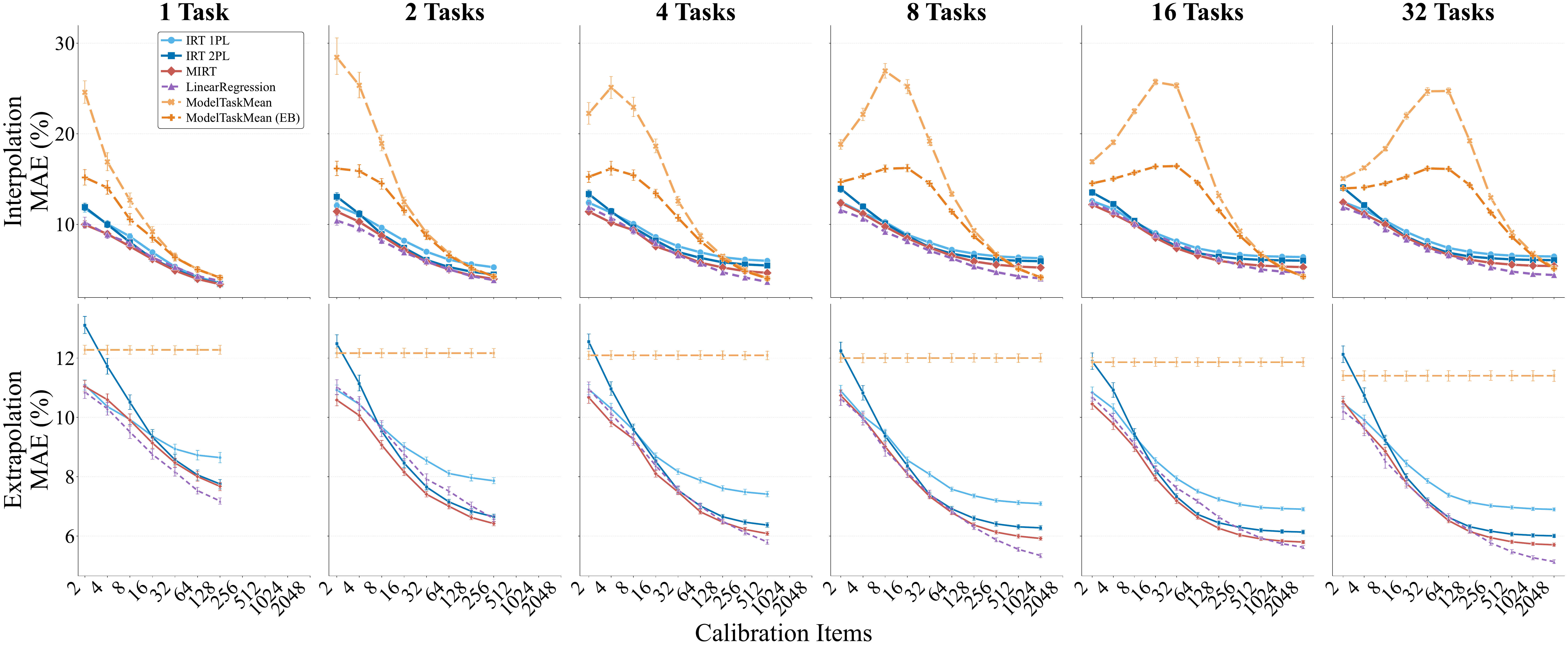}
  \caption{Full comparison of the (Nested) MIRT model against the IRT and Model Task Mean predictors. Model Task Mean (EB) represents the empirical bayes version of the estimator. Assessors receive random samples from $T$ calibration tasks (interpolation) and then must predict how the model will perform on those $T$ calibration tasks and on $128 - T$ validation tasks (extrapolation). The MIRT model validation MAE drops both when the number of calibration tasks or the number of samples increases.}
  \label{fig:calib_valid_full}
\end{figure*}

\paragraph{Interpolation Results}
\Cref{fig:calib_valid_full} presents the full grid of interpolation and extrapolation results for the prediction experiments. Note that the second row here focused on extrapolation results is partially redundant with \Cref{fig:mirt_baselines_and_prediction} in the main body of this work, but is given again here for reference.
In \Cref{tab:interpolation_mae} we present interpolation results for the single task case.
This reflects the setup used in prior work where a single benchmark is sampled, and the task is to compress it to a small set of items.
In this setting, our MIRT implementation with VOptimal selection tends to still outperform all other methods with few exceptions.

\begin{table}[t]
  \centering
  \caption{Interpolation MAE (\%) for $n_{\text{tasks}} = 1$. Each method uses its best available selector. Lower is better; \textbf{bold} marks the best method per column. This setting simulates prior work that focused on compressing a single benchmark.}
  \label{tab:interpolation_mae}
  \begin{tabular}{lrrrrrrr}
    \toprule
     &  &  &  &  \textbf{Items} &  &  & \\
    \textbf{Method} & \textbf{2} & \textbf{4} & \textbf{8} & \textbf{16} & \textbf{32} & \textbf{64} & \textbf{128} \\
    \midrule
    IRT 1PL & 11.76 & 10.03 & 8.65 & 6.91 & 5.33 & 4.19 & 3.52 \\
    IRT 2PL + Fisher & 8.93 & 7.68 & 6.48 & \textbf{5.39} & 4.67 & \textbf{3.90} & 3.41 \\
    MIRT + VOptimal & \textbf{8.53} & \textbf{7.43} & \textbf{6.42} & 5.46 & \textbf{4.58} & 3.94 & \textbf{3.38} \\
    Linear Regression & 10.20 & 8.86 & 7.90 & 6.33 & 5.29 & 4.29 & 3.74 \\
    ModelTaskMean (EB) & 15.17 & 14.05 & 10.55 & 8.52 & 6.35 & 5.01 & 4.11 \\
    \bottomrule
  \end{tabular}
\end{table}

\subsection{Factor Analysis}

\label{sec:factor_analysis}

\begin{figure}[htbp]
    \centering
    \begin{minipage}{0.48\textwidth}
        \centering
        \includegraphics[width=\textwidth]{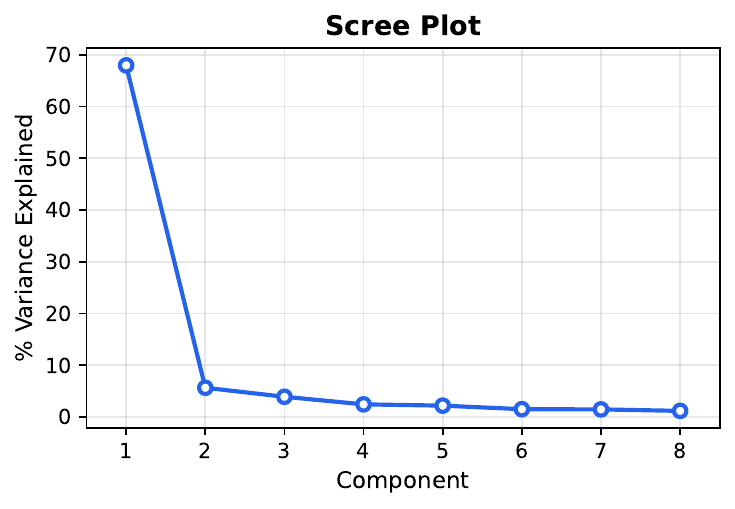}
    \end{minipage}
    \hfill
    \begin{minipage}{0.48\textwidth}
        \centering
        \includegraphics[width=\textwidth]{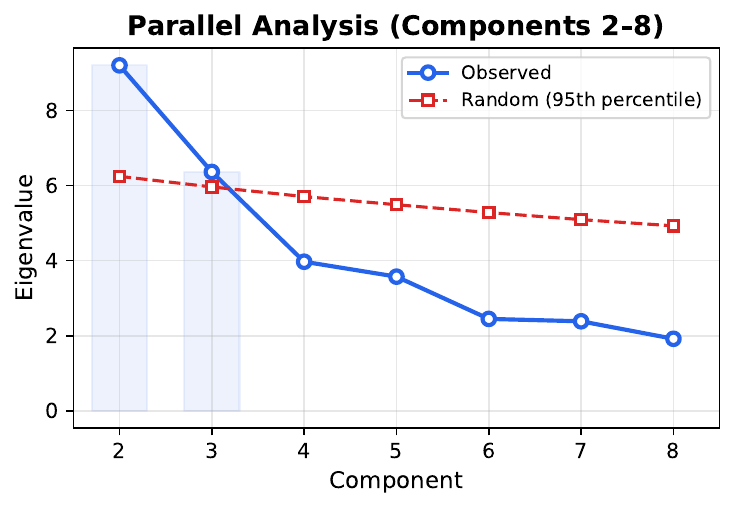}
    \end{minipage}
    \caption{Scree plot and parallel analysis showing the most important factors in the training response matrix. A single factor explains almost 70\% of variance, and two more latent factors can be selected via the elbow method.}
    \label{fig:combined}
\end{figure}
We perform a factor analysis of the WILD dataset. \Cref{fig:combined} presents a scree plot (left) of principal components, and a parallel analysis to visualize the first 8 eigenvalues (right) starting at the second, in order to zoom in on the factors $>$1.
The scree plot shows that one factor explains almost 70\% of the variance, consistent with one general ability, or ``g-factor" across benchmarks.
The parallel analysis shows that up to 3 components exceed the 95th percentile eigenvalue threshold of random data of equivalent dimensions. 
Given this, we set $K$=3 for all MIRT experiments in the main results.
We also ablate $K$ in further analysis (\Cref{fig:k_ablation}), which shows that in some cases either 2, or more than 3 dimensions may attain stronger performance in the prediction setting.
However, we maintain $K$=3 as to not pollute the main results from observations on the test set.

\begin{figure}[t]
  \centering
  \includegraphics[width=0.5\linewidth]{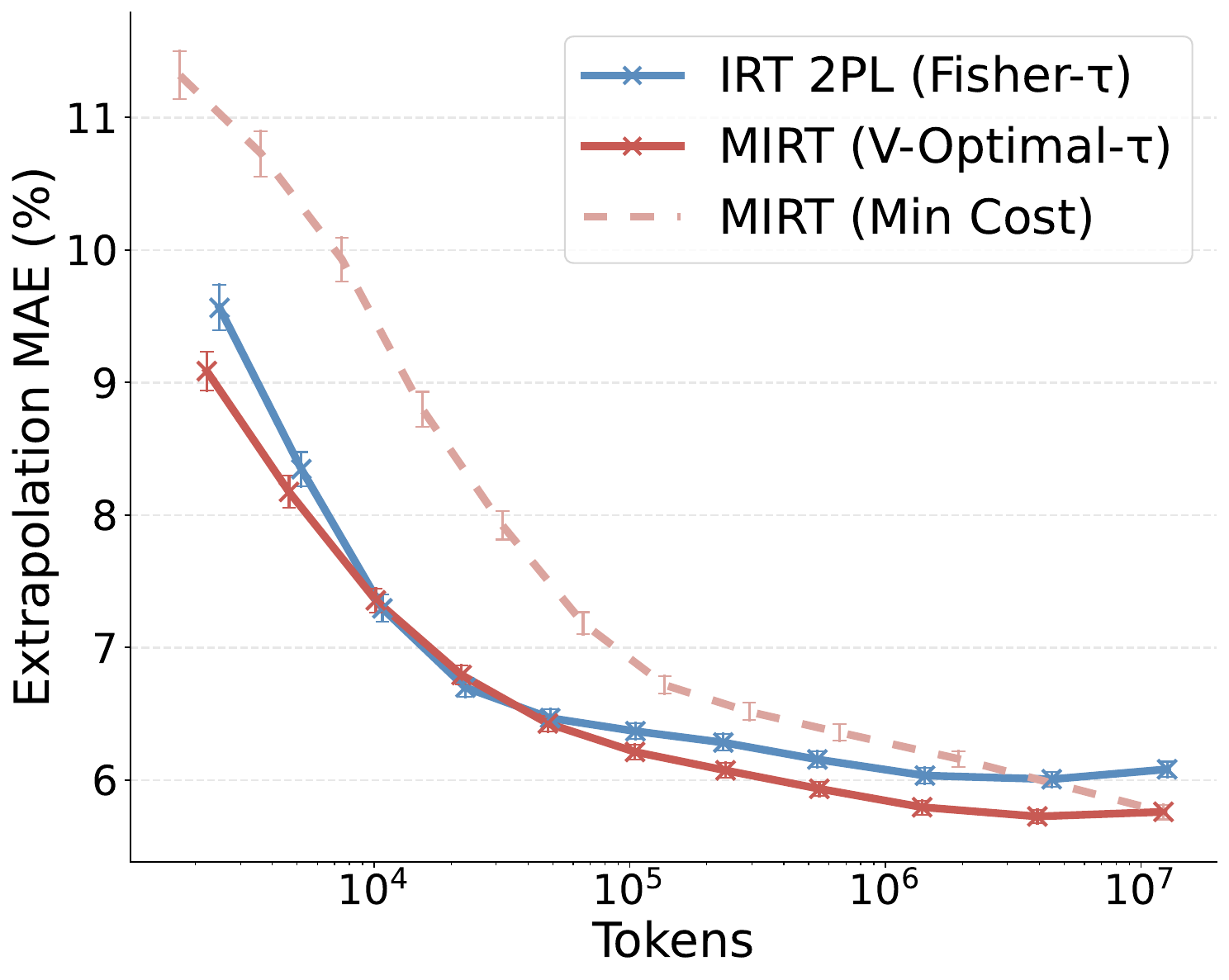}
  \caption{Comparison of cost discounted models to the minimum cost baseline. All results presented for 16 calibration tasks.}
  \label{fig:min_cost}
\end{figure}
\section{Minimum Cost Selection}
\label{sec:min_cost_selection}
\Cref{fig:min_cost} compares the token-aware discounted models against minimum cost baseline selectors, where the cheapest items in terms of token cost are greedily selected.
The discounted models outperform the minimum cost models by a large margin, which shows that these cheap items are not informative enough for capability estimation, and information-aware selection is crucial.

\section{Reasoning Model Preliminary Study}\label{ref:reasoning_study}

\begin{figure}
    \centering
    \includegraphics[width=\linewidth]{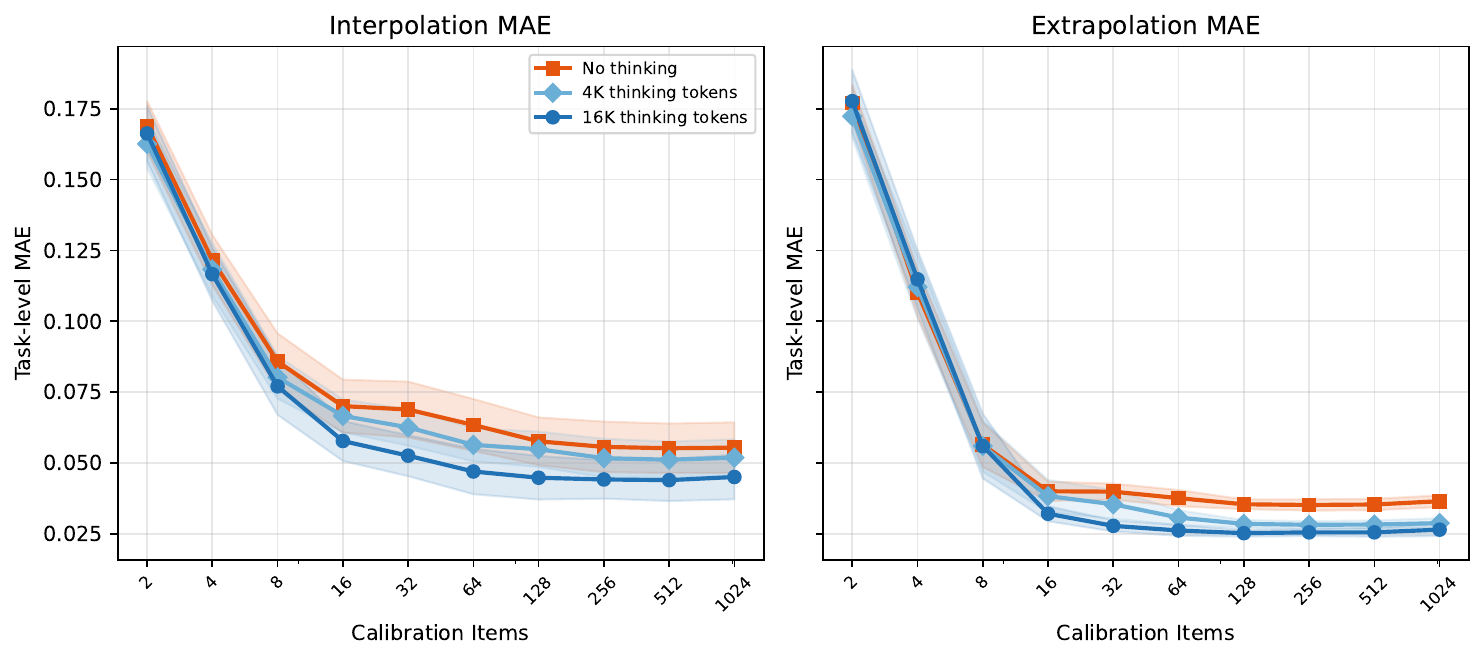}
    \caption{Analysis of interpolation and extrapolation prediction performance for Sonnet 4.6 at different reasoning budgets. Our method is robust to the model behavior changes related to newer reasoning abilities with large thinking budgets.}
    \label{fig:reasoning_model}
\end{figure}

  \begin{table}[t]
  \centering
  \begin{tabular}{ll}
  \toprule
  \textbf{Calibration Tasks} & \textbf{Held-out Tasks} \\
  \midrule
  BoolQ              & BBH (Object Counting) \\
  ChemBench (Organic Chemistry) & BigCodeBench \\
  DROP               & GSM8K \\
  HellaSwag          & MedQA \\
  MATH (Precalculus) & MMLU-Pro (Health) \\
  PAWS               & MuSR \\
  SQuAD              & PiQA \\
  TruthfulQA         & RACE-H \\
  \bottomrule
  \end{tabular}
  \caption{Tasks used for Sonnet 4.6 evaluation.}
  \label{tab:eval-tasks}
  \end{table}

To evaluate the potential impact of extending the presented approach to reasoning models, we ran a preliminary experiment with Sonnet 4.6, a reasoning model not contained within WILD. We evaluated Sonnet on samples from 16 tasks. We used the same evaluation setting described in \Cref{ref:experimental_setup}, with 8 of the tasks selected as calibration tasks and the remaining 8 held out as validation tasks. The full list of tasks is in \Cref{tab:eval-tasks}. We ran the MIRT model described in the paper with the VOptimal selection in the adaptive setting. The results are shown in \Cref{fig:reasoning_model}. We did not observe any significant degradation in performance when the model is run with reasoning. However, we leave a more thorough investigation of the predictability of reasoning models for future work.

\section{MIRT Ablations}
\label{sec:MIRT ablations}
\begin{figure}[h]
    \centering
    \includegraphics[width=\linewidth]{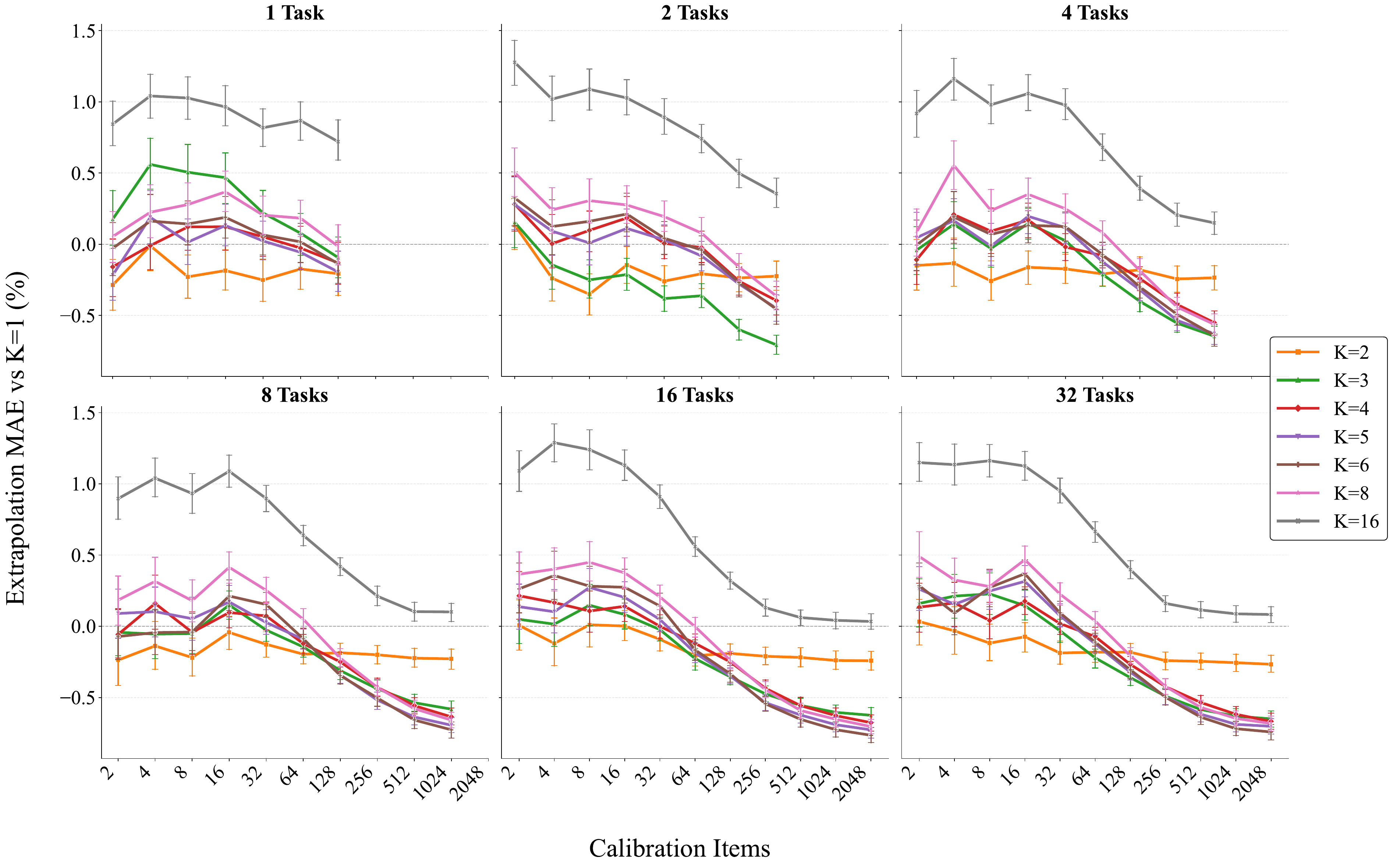}
    \caption{Ablation over $K$ value for the MIRT model. Extrapolation MAE as a \% difference from $K$=1. These results reflect the vanilla MIRT model, not Nested MIRT.}
    \label{fig:k_ablation}
\end{figure}

\begin{figure}[h]
    \centering
    \includegraphics[width=\linewidth]{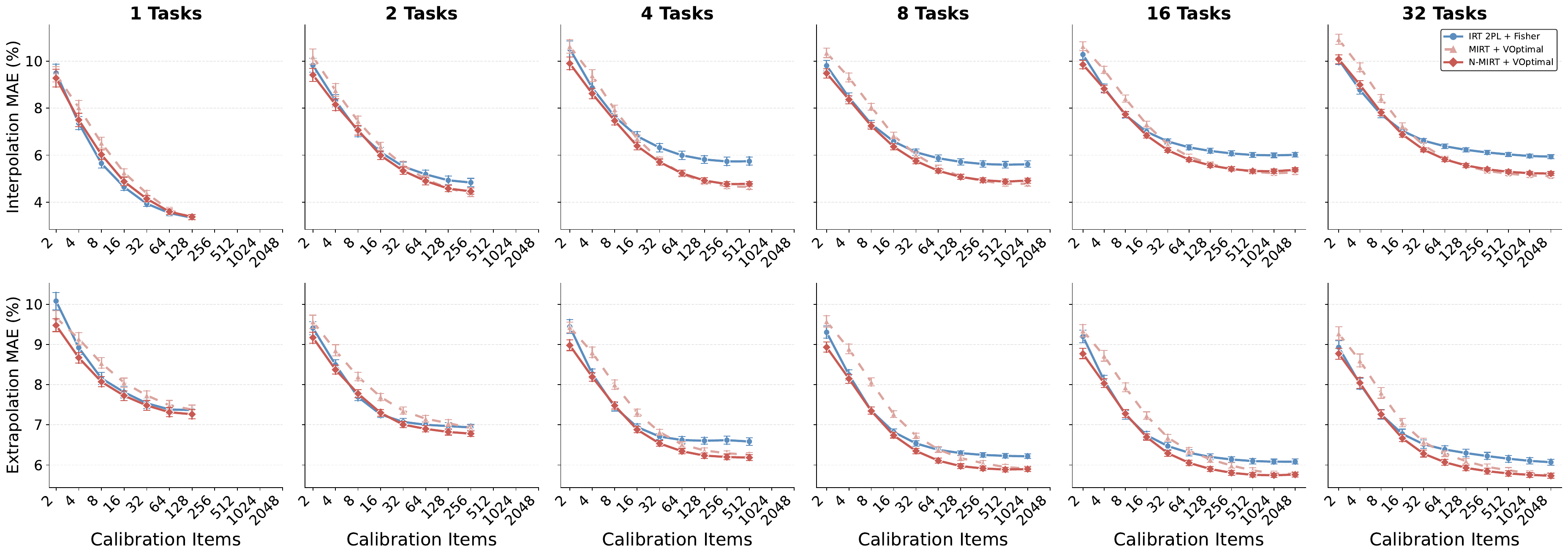}
    \caption{Comparison of MIRT with the Nested MIRT model for adaptive experiments. We plot the IRT 2PL model as a baseline comparison. Both MIRT models have $K$=3 parameters (including the general loading).}
    \label{fig:nested_v_mirt_ablation}
\end{figure}
In \Cref{fig:k_ablation}, we show the impact of varying $K$ at different numbers of calibration tasks and items.
Surprisingly, for 1 task, setting $K$=3 as we do in our main experiments is rather weak, though it is a reasonable hyperparameter as we increase the number of tasks and items.
We also find that higher $K$ is a strong choice at higher numbers of tasks and items, likely due to its higher representational capacity, and the fact that with sufficient items overfitting is not a concern.
$K$=2 is a strong choice for low numbers of tasks and items, but is a weak model as more items are sampled.
In \Cref{fig:nested_v_mirt_ablation} we show that the adaptive IRT 2PL model often outperforms MIRT at low numbers of calibration tasks and items.
The Nested MIRT model closes this gap while maintaining the MIRT model's strong performance as items and tasks increase.
\begin{figure}[t]
    \centering
    \includegraphics[width=0.8\linewidth]{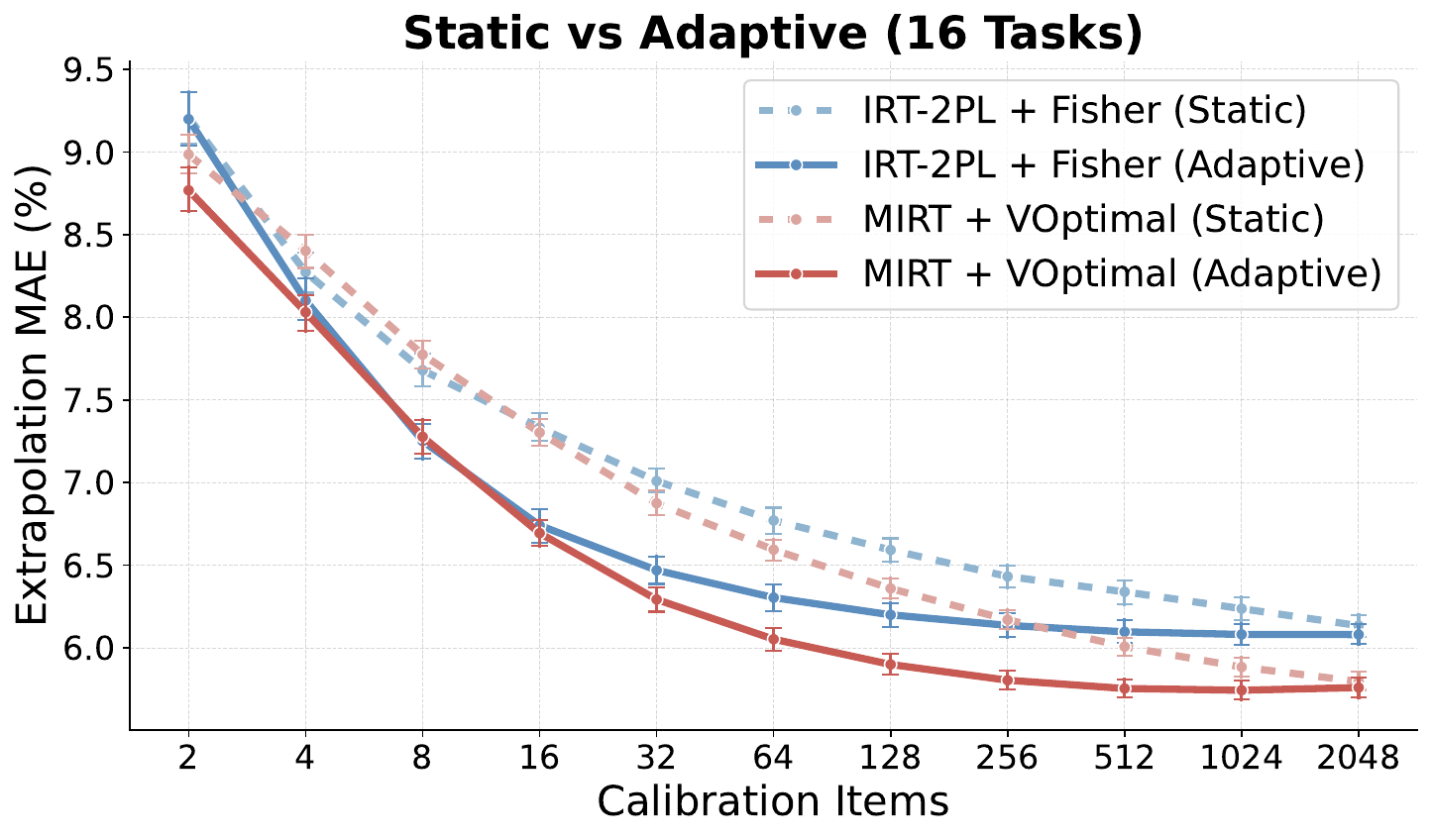}
    \caption{Comparison of adaptive and static selection on the same plot.}
    \label{fig:adaptive_v_static}
\end{figure}

\begin{figure}
  \centering
  \includegraphics[width=0.7\columnwidth]{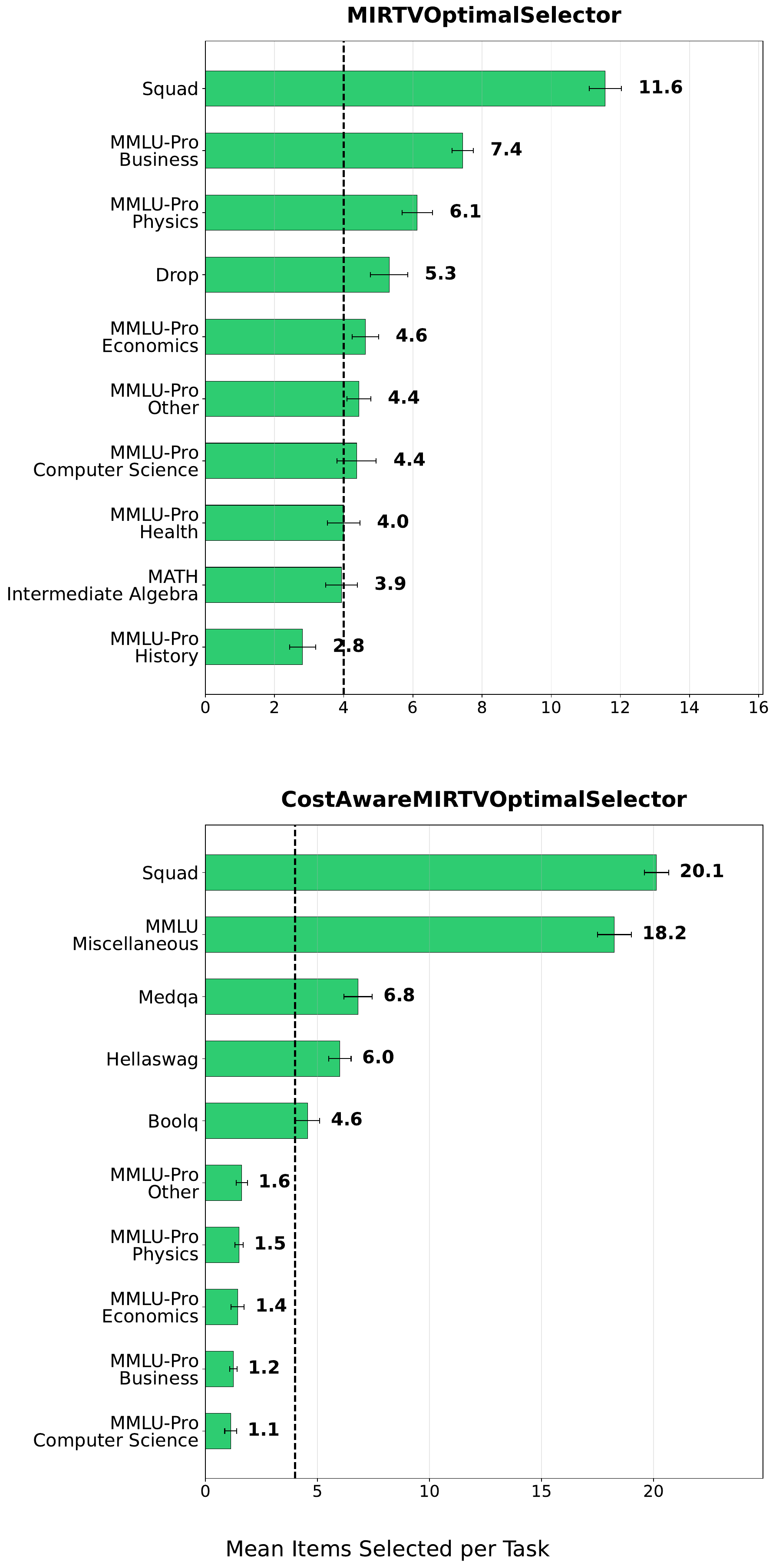}
  \caption{Comparison of the task selection rates between the V-Optimal and cost-aware V-Optimal (V-Optimal-$\tau$) when selecting 64 calibration items. Dotted line represents expected amount of occurrences if the selection was random.}
  \label{fig:task_selection_plot}
\end{figure}
\section{Selection Analysis}
\label{sec:select-analysis}

We analyze the rate at which the VOptimal and VOptimal-$\tau$ select tasks. We compute the average number of items selected per task, given the task is one of the calibration tasks. 
\Cref{fig:task_selection_plot} shows the top 10 tasks by this metric when there are 64 items to select. VOptimal seems to have a strong preference for Squad and various MMLU-Pro tasks. The cost-aware version focuses on similar MMLU-Pro tasks at lower rates, but heavily up-samples Squad and MMLU Miscellaneous. We hypothesize that the bias towards MMLU Miscellaneous is driven mainly by its low token cost, with an average of 120 total tokens per answer.
Other low-cost tasks like ARC, Commonsense QA, and PAWS are noticeably absent, indicating that they were not as highly informative according to our model.
MATH Intermediate Algebra and DROP, relatively costly tasks, are frequently sampled by the VOptimal model but are absent in the cost aware selection which samples from cheaper QA tasks and hellaswag.

\section{Limitations}
This work has a few potential limitations. First, we model only binary correctness labels, ignoring richer response information such as partial credit, answer quality, or confidence scores that could provide more nuanced ability estimates. IRT models can certainly be fit on non-binary data, but we focus our results on the binary case only. Second, our benchmark consists primarily of short-horizon tasks with clear correct answers. Long-horizon tasks (e.g., multi-step planning, extended dialog, or complex tool use) may require abilities not well-represented in our current item pool, potentially limiting the generalizability of our ability estimates to such settings.

\begin{figure}[t]
    \centering
    \includegraphics[width=0.85\textwidth]{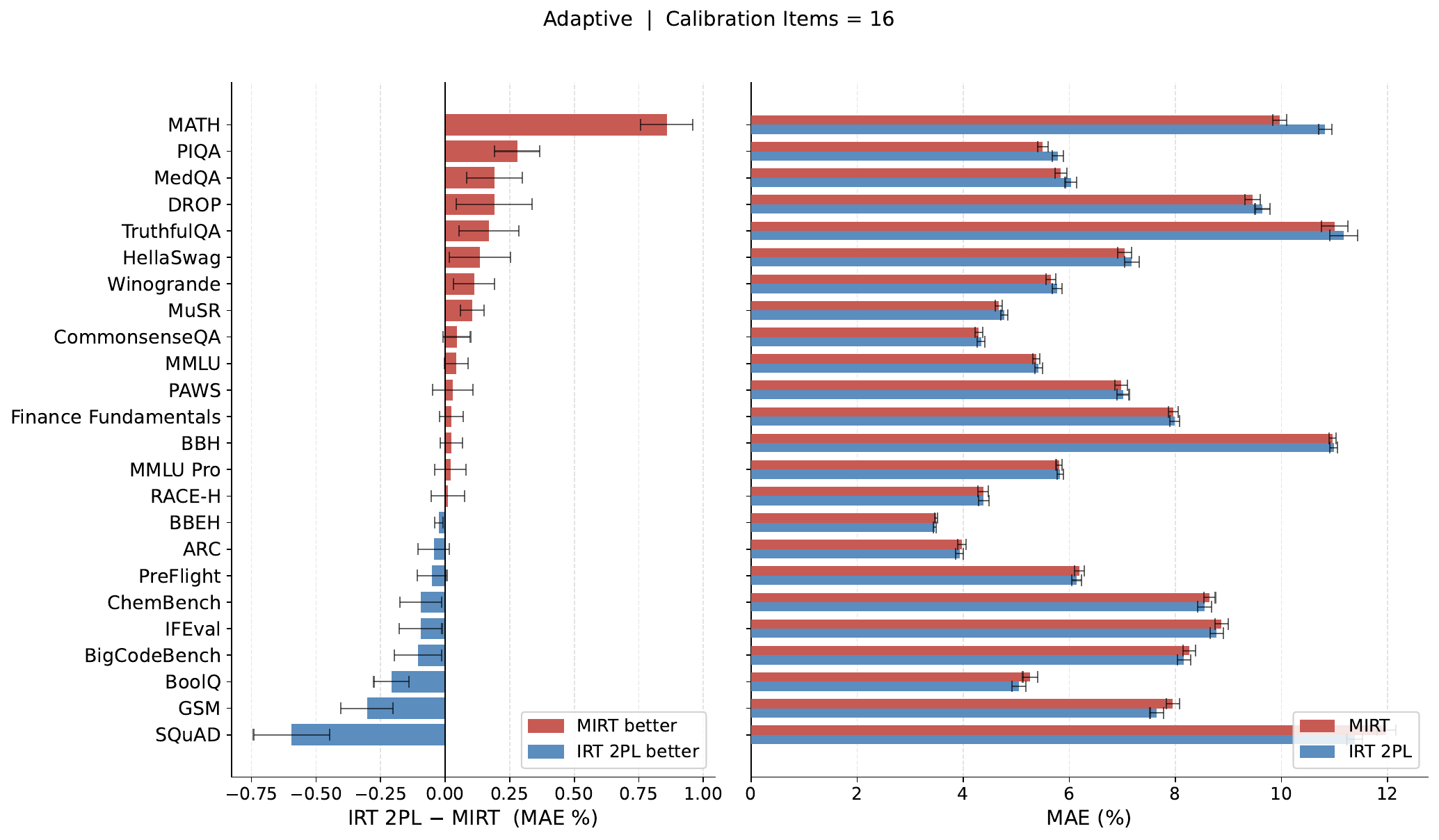}
    \caption{Results breakdown per-task for adaptive models with 16 calibration tasks and 16 calibration items.
    \textbf{Left}: Relative MAE comparison of the IRT-2PL model with Fisher selection vs. MIRT with VOptimal selection. Further left in blue means that the 2PL model attained a lower MAE, whereas further right in red means that the MIRT model had a lower MAE.
    \textbf{Right}: Absolute MAE comparisons per task. Lower is better.}
    \label{fig:per_task_n_16}
\end{figure}
\begin{figure}[t]
    \centering
    \includegraphics[width=0.85\textwidth]{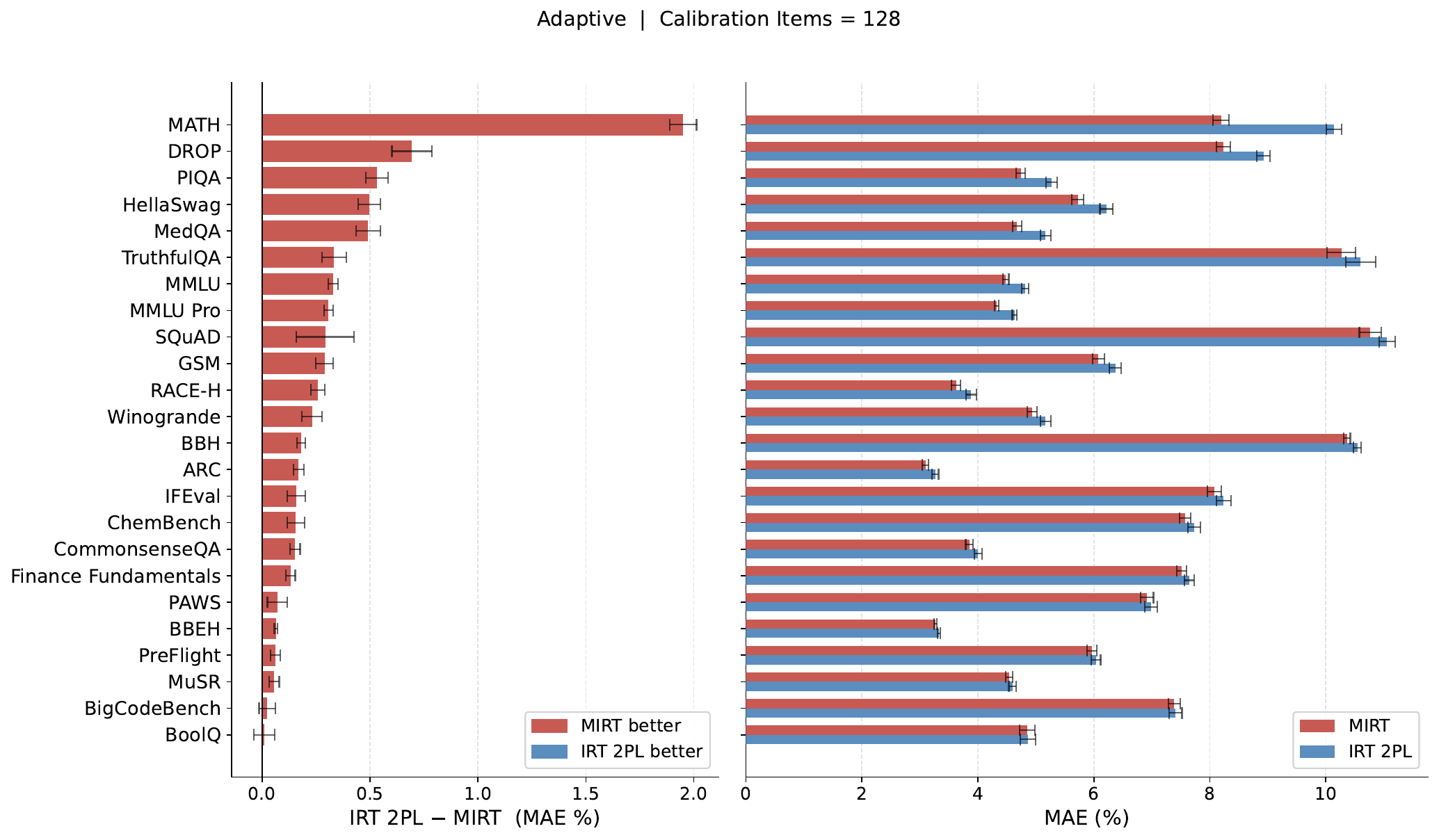}
    \caption{Results breakdown per-task for adaptive models with 16 calibration tasks and 128 calibration items.
    \textbf{Left}: Relative MAE comparison of the IRT-2PL model with Fisher selection vs. MIRT with VOptimal selection. Further left in blue means that the 2PL model attained a lower MAE, whereas further right in red means that the MIRT model had a lower MAE.
    \textbf{Right}: Absolute MAE comparisons per task. Lower is better.}
    \label{fig:per_task_n_128}
\end{figure}

\begin{figure}[t]
    \centering
    \includegraphics[width=0.85\textwidth]{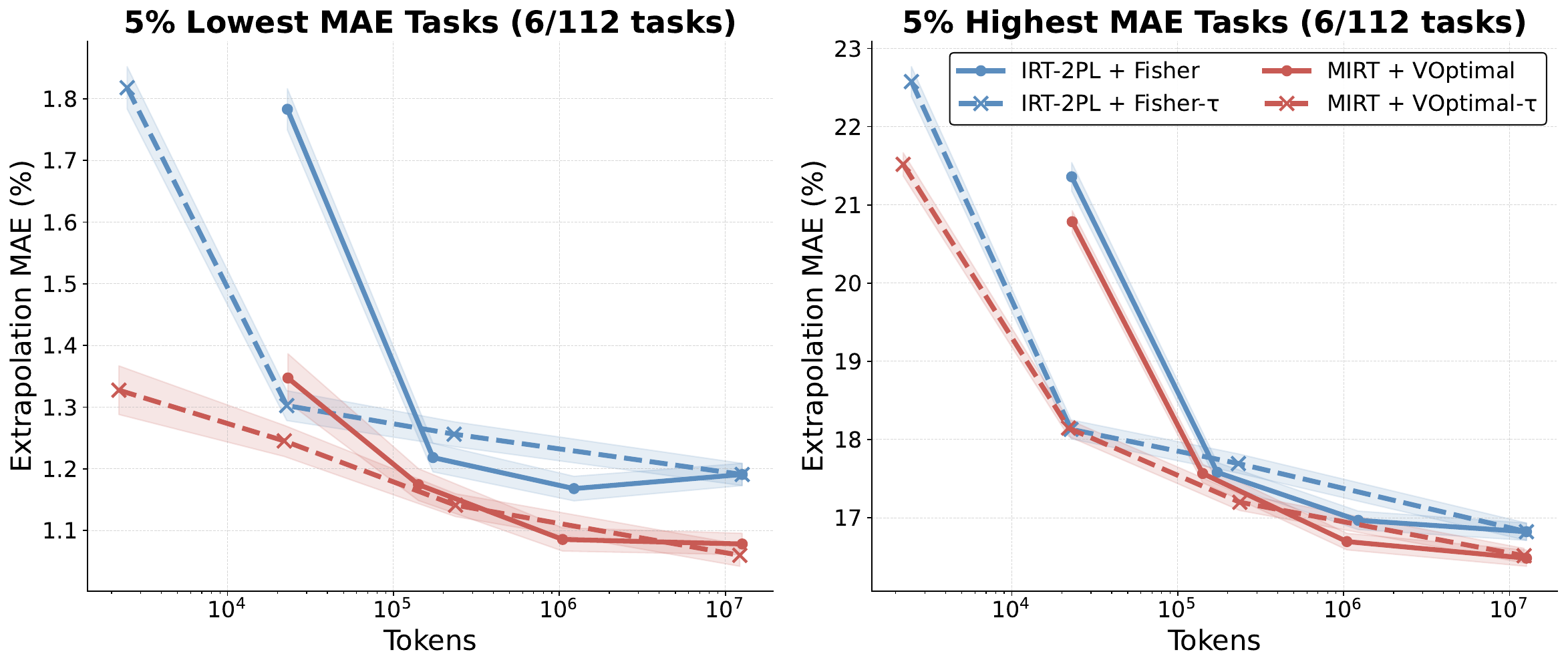}
    \caption{Comparison of assessor performance for the 5\% best-performing, i.e., lowest MAE tasks (left) and 5\% worst-performing, i.e., highest MAE tasks (right).}
    \label{fig:best_worst_task_mae}
\end{figure}

\begin{figure}[t]
    \centering
    \includegraphics[width=0.85\textwidth]{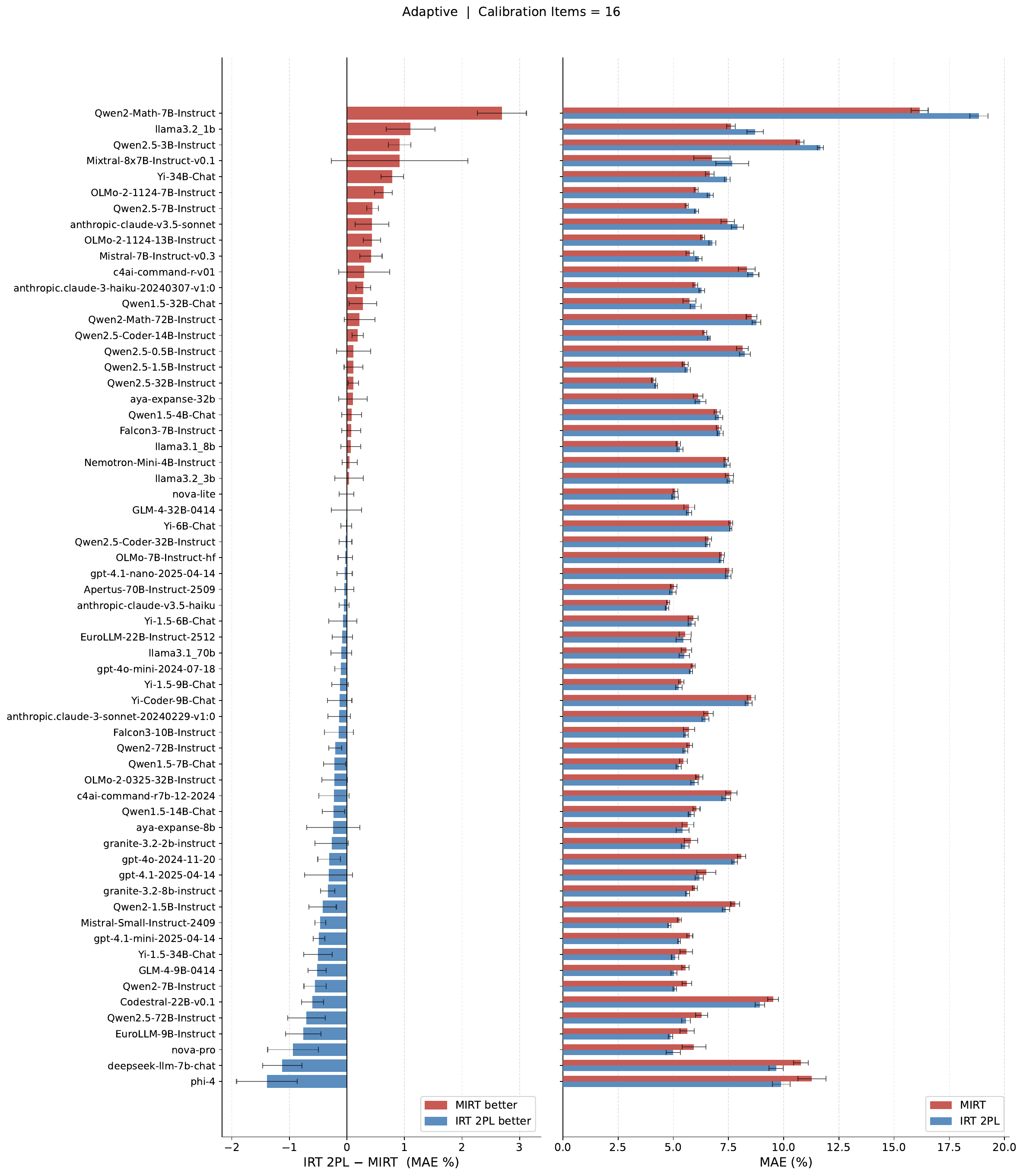}
    \caption{Results breakdown per-model for adaptive models with 16 calibration tasks and 16 calibration items.
    \textbf{Left}: Relative MAE comparison of the IRT-2PL model with Fisher selection vs. MIRT with VOptimal selection. Further left in blue means that the 2PL model attained a lower MAE, whereas further right in red means that the MIRT model had a lower MAE.
    \textbf{Right}: Absolute MAE comparisons per task. Lower is better.}
    \label{fig:per_model_n_16}
\end{figure}
\begin{figure}[t]
    \centering
    \includegraphics[width=0.85\textwidth]{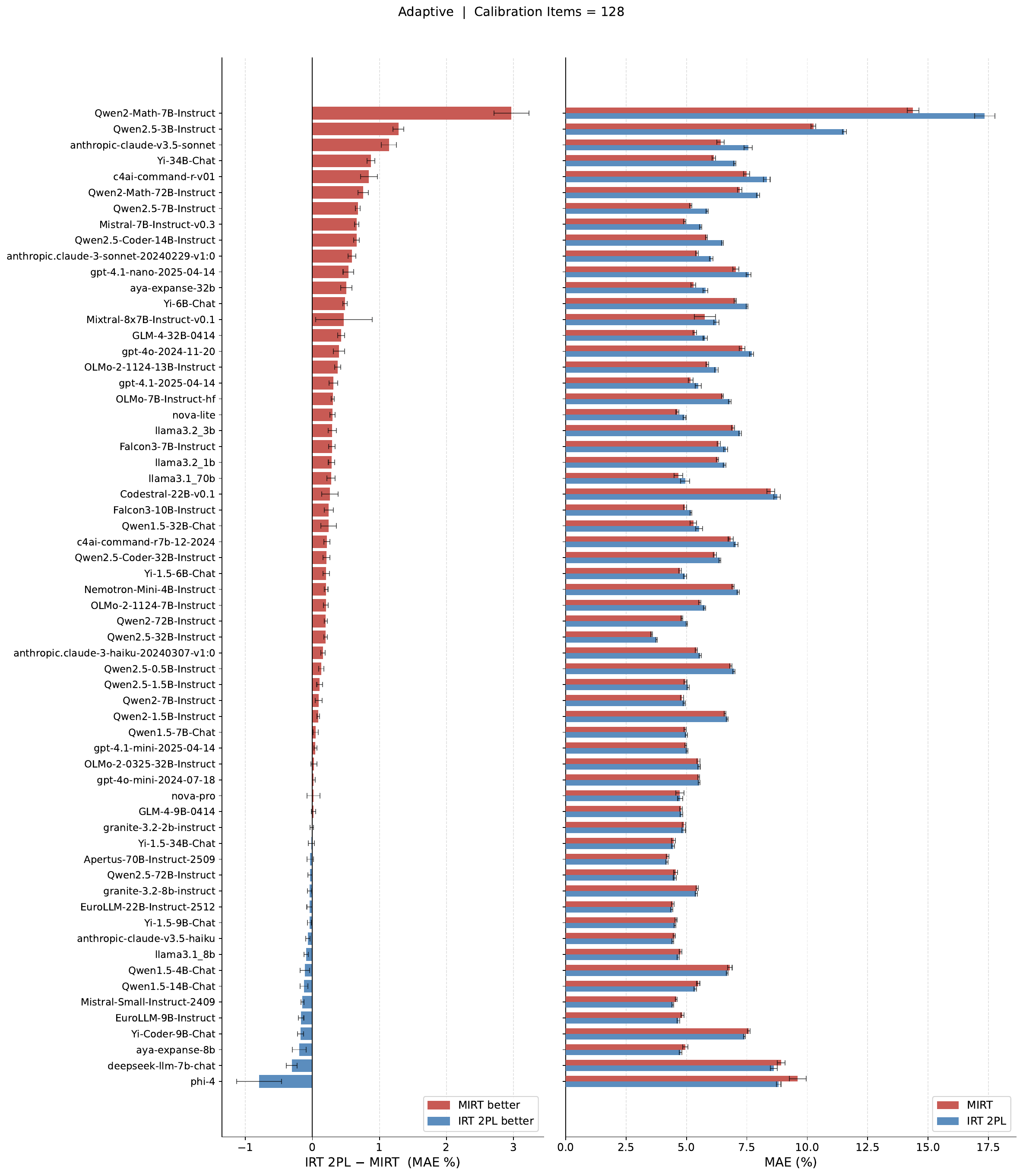}
    \caption{Results breakdown per-model for adaptive models with 16 calibration tasks and 128 calibration items.
    \textbf{Left}: Relative MAE comparison of the IRT-2PL model with Fisher selection vs. MIRT with VOptimal selection. Further left in blue means that the 2PL model attained a lower MAE, whereas further right in red means that the MIRT model had a lower MAE.
    \textbf{Right}: Absolute MAE comparisons per task. Lower is better.}
    \label{fig:per_model_n_128}
\end{figure}
\section{Further Results Analysis: Task and Model Variance}
\label{sec:results_variance}
We compare the IRT-2PL with Fisher selection assessor to the MIRT with VOptimal selection assessor at the task level in \Cref{fig:per_task_n_16} and \Cref{fig:per_task_n_128}, where results are averaged over LLMs.
At 16 calibration items (\Cref{fig:per_task_n_16}), where both models perform similarly, we see a split in which model is best for which task.
MIRT is much stronger for MATH, whereas IRT-2PL is stronger for SQUAD. 
Both of these tasks are among the highest MAE for both models. 
Big Bench Hard (BBH) is very high MAE at approximately the same performance for both models.
At 128 calibration items (\Cref{fig:per_task_n_128}), MIRT attains a lower MAE on every task, and the gap in MATH increases.
SQUAD remains notably difficult to predict, whereas HellaSwag, for instance, has a large drop in MAE.
\Cref{fig:best_worst_task_mae} shows that as we sample more tasks, adaptive MIRT consistently outperforms the 2PL model on its best performing tasks, and, with a smaller gap in MAE, on its worse performing tasks too.

In \Cref{fig:per_model_n_16} and \Cref{fig:per_model_n_128} we compare these assessors at the model level, where results are averaged over tasks.
At 16 calibration items, both assessors perform similarly for several models, with a slightly longer tail of models that IRT-2PL performs best on.
The MIRT assessor has a large drop in MAE for Qwen2-MATH-7b-instruct---the most difficult model to predict---relative to the IRT-2PL assessor.
At 128 calibration items, the MIRT assessor is best for almost all models.
One exception to this is phi-4, for which IRT-2PL maintains a much smaller MAE than MIRT.

\section{Held Out Estimation}
\label{sec:held-out}

The goal of our analysis is evaluating how well our assessors can predict $\theta_{m, t}$, the true probability of model $m$ correctly answering an item from task $t$. This is distinct from evaluating how well the model can predict $\mu_r(\mathcal{R}_t)$. 

For each task $t$, we can partition the total set of records $\mathcal{R}_t$ for that task into two sets, $\mathcal{O}_t$ representing the observed items for the assessor and $\mathcal{U}_t$ the complementary unobserved items s.t. \(\mathcal{O}_t \cup \mathcal{U}_t = \mathcal{R}_t\) and \(\mathcal{O}_t \cap \mathcal{U}_t = \emptyset\). We do not constrain the observed set to be of equal size across tasks (e.g. \(|\mathcal{O}_1| \ne |\mathcal{O}_2|\)), but we do enforce that any sampling algorithm may only be a function of \(\mathcal{O}_t\).

$\theta_{m, t}$ is a latent variable we can never directly observe. In order to measure the quality of the assessor we need a proxy to use as ground truth. It is common for existing work on data set compression to use the mean over the entire set of available data,
\(\mu_r(\mathcal{R}_t)\). In this work, we take special care to always use a separate
set of data unobserved by the assessor, $\mu_r(\mathcal{U}_t)$.

For an arbitrary prediction \(q_{m,t}\) that is a function of \(\mathcal{O}_t\), we can write for
the held-out set,
\begin{equation}
\mathbb{E} \left[ (q_{m,t} - \mu_r(U_t) )^2 \right] = \mathbb{E}[(q_{m, t} - \theta_{m, t})^2] + \text{Var}(\mu_r(U_t)).
\end{equation}

Whereas for the total set, the error decomposes into three terms,
\begin{equation}
\mathbb{E} \left[ (q_{m,t} - \mu_r(\mathcal{R}_t) )^2 \right] = \mathbb{E}[(q_{m,t} - \theta_{m, t})^2] + \text{Var}(\mu_r(\mathcal{R}_t)) - 2 \frac{|\mathcal{O}_t|}{| \mathcal{R}_t | } \text{Cov}(q_{m,t}, \mu_r(\mathcal{O}_t)).
\end{equation}

With the held-out target, every assessor's error decomposes into its squared
error with the true \(\theta_{m, t}\), plus a per-task sampling noise constant. Since this
constant only varies by task, assessor performance can be compared directly
without controlling for the size of the observed set.

With the full-set target, the error contains an additional assessor-specific
covariance term to account for the fact the observed items are used to both form
a prediction and to compute the target itself. This particularly confounds
comparisons between assessors who have received disparate, non-uniform item
samples across the tasks.

When the proxy is \(\mu_r(\mathcal{U}_t)\), \(\theta_{m, t}\) is the optimal prediction,
\begin{equation}
\mathbb{E} \left[ (\theta_{m,t} - \mu_r({\mathcal{U}_t} 
))^2 \right] = \text{Var}(\mu_r(\mathcal{U}_t)) = \frac{\theta_{m, t} (1 - \theta_{m, t})}{|\mathcal{U}_t|}.
\end{equation}

However, when the proxy is instead \(\mu_r(\mathcal{R}_t)\), we can improve the predictor by
smoothing it with the observed items. The expected error is,
\begin{equation}
\mathbb{E} \left[ (\theta_{m,t} - \mu_r({\mathcal{R}_t} 
))^2 \right] = \text{Var}(\mu_r(\mathcal{R}_t)) = \frac{\theta_{m, t} (1 - \theta_{m, t})}{|\mathcal{R}_t|}.
\end{equation}
We can instead use
\begin{equation}
\tilde{\theta}_{m, t} = \frac{|\mathcal{O}_t|}{|\mathcal{R}_t|} \mu_r({\mathcal{O}_t}) + \frac{|\mathcal{U}_t|}{|\mathcal{R}_t|} \theta_{m, t}.
\end{equation}
This gives us a lower expected error of
\begin{equation}
\mathbb{E} \left[ (\tilde{\theta}_{m, t} - \mu_r({\mathcal{R}_t} ))^2 \right] = \text{Var}(\mu_r({\mathcal{R}_t})) \left(1 - \frac{|\mathcal{O}_t|}{|\mathcal{R}_t|} \right).
\end{equation}

Since our goal is recovering \(\theta_{m, t}\) rather than directly compressing the exact
benchmark scores, we grade only against the held-out targets.

\begin{table}
\centering
\small
\caption{Data partitioning schema for calibration and validation.}
\label{tab:data_partitioning}
\begin{tabular}{@{}llll@{}}
\toprule
\textbf{Category} & \textbf{Models} & \textbf{Tasks} & \textbf{Access} \\
\midrule
Training     & $\mathcal{M}_{train}$ & $\mathcal{T}^c \cup \mathcal{T}^v$ & Full obs. \\
Calibration  & $\mathcal{M}_{test}$ & $\mathcal{T}^c$ (item subset)   & Test-time \\
Interpolation & $\mathcal{M}_{test}$ & $\mathcal{T}^c$ (held-out items) & \textbf{Predicted} \\
Extrapolation & $\mathcal{M}_{test}$ & $\mathcal{T}^v$            & \textbf{Predicted} \\
\bottomrule
\end{tabular}
\end{table}

\section{Model \& Task List}
\label{sec:detailed-model-task}
\pagebreak
\begin{table*}[ht]
\centering
\caption{Benchmarks, tasks, and number of unique evaluation items contained in WILD}
\label{tab:benchmarks_unique_items}
\scriptsize
\begin{tabular}{lcr}
\toprule
\textbf{Dataset} & \textbf{Task} & \textbf{Items} \\
\midrule

WinoGrande \cite{10.1145/3474381} &  & 1,267 \\
TruthfulQA \cite{lin-etal-2022-truthfulqa} &  & 817 \\
SQuAD \cite{rajpurkar-etal-2016-squad} &  & 11,866 \\
RACE-H \cite{lai-etal-2017-race} &  & 3,498 \\
Pre-Flight \cite{inspect_evals_preflight} &  & 300 \\
PIQA \cite{Bisk2020} &  & 1,803 \\
PAWS \cite{paws2019naacl} &  & 7,982 \\
MUSR \cite{DBLP:conf/iclr/SpragueYBCD24} &  & 250 \\
DROP \cite{dua-etal-2019-drop} &  & 8,687 \\
BoolQ \cite{clark-etal-2019-boolq} &  & 3,270 \\
BigCodeBench \cite{zhuo2024bigcodebench} &  & 1,139 \\
\midrule
\multirow{25}{*}{\textbf{BBH} \cite{suzgun-etal-2023-challenging}}
 & Boolean Expressions & 250 \\
 & Date Understanding & 250 \\
 & Disambiguation QA & 250 \\
 & Dyck Languages & 250 \\
 & Formal Fallacies & 250 \\
 & Geometric Shapes & 250 \\
 & Multistep Arithmetic (Two) & 250 \\
 & Navigate & 250 \\
 & Object Counting & 250 \\
 & Reasoning About Colored Objects & 250 \\
 & Salient Translation Error Detection & 250 \\
 & Temporal Sequences & 250 \\
 & Tracking Shuffled Objects (3) & 250 \\
 & Tracking Shuffled Objects (5) & 250 \\
 & Tracking Shuffled Objects (7) & 250 \\
 & Web of Lies & 250 \\
 & Word Sorting & 250 \\
 & Movie Recommendation & 249 \\
 & Sports Understanding & 248 \\
 & Ruin Names & 210 \\
 & Causal Judgement & 185 \\
 & Penguins in a Table & 146 \\
 & Logical Deduction (3) & 99 \\
 & Logical Deduction (5) & 98 \\
 & Logical Deduction (7) & 95 \\
\midrule
\multirow{23}{*}{\textbf{BBEH} \cite{kazemi-etal-2025-big}}
 & Boardgame QA & 200 \\
 & Boolean Expressions & 200 \\
 & Buggy Tables & 200 \\
 & Causal Understanding & 200 \\
 & Dyck Languages & 200 \\
 & Geometric Shapes & 200 \\
 & Hyperbaton & 200 \\
 & Movie Recommendation & 200 \\
 & Multistep Arithmetic & 200 \\
 & NYCC & 200 \\
 & Object Counting & 200 \\
 & Object Properties & 200 \\
 & Sarcasm Triples & 200 \\
 & Shuffled Objects & 200 \\
 & SportQA & 200 \\
 & Temporal Sequence & 200 \\
 & Time Arithmetic & 200 \\
 & Web of Lies & 200 \\
 & Word Sorting & 200 \\
 & Zebra Puzzles & 200 \\
 & Spatial Reasoning & 199 \\
 & Linguini & 194 \\
 & Disambiguation QA & 120 \\

\midrule

\multirow{2}{*}{ARC \cite{DBLP:journals/corr/abs-1803-05457}} 
& Easy & 2,371 \\
& Challenge &  1,170 \\
\midrule

\multirow{2}{*}{AIME \cite{aime_maa}}
 & 2024 & 30 \\
 & 2025 & 30 \\
\midrule

\multirow{14}{*}{\textbf{MMLU Pro} \cite{wang2024mmlu}} 
 & Math & 1,349 \\
 & Physics & 1,294 \\
 & Chemistry & 1,132 \\
 & Law & 1,023 \\
 & Engineering & 968 \\
 & Other & 924 \\
 & Economics & 840 \\
 & Psychology & 796 \\
 & Business & 779 \\
 & Health & 778 \\
 & Biology & 634 \\
 & Philosophy & 498 \\
 & Computer Science & 409 \\
 & History & 381 \\
\midrule

\bottomrule
\end{tabular}
\end{table*}

\begin{table*}
\centering
\scriptsize
\begin{tabular}{lcc}
\toprule
\textbf{Dataset} & \textbf{Task} & \textbf{Items} \\
\midrule
\multirow{57}{*}{\textbf{MMLU} \cite{hendrycks2021measuring}}
 & Professional Law & 1,534 \\
 & Moral Scenarios & 895 \\
 & Miscellaneous & 783 \\
 & Professional Psychology & 609 \\
 & High School Psychology & 534 \\
 & High School Macroeconomics & 386 \\
 & Elementary Mathematics & 376 \\
 & Moral Disputes & 346 \\
 & Prehistory & 323 \\
 & Philosophy & 311 \\
 & High School Biology & 310 \\
 & Nutrition & 299 \\
 & Professional Accounting & 281 \\
 & Professional Medicine & 272 \\
 & High School Mathematics & 265 \\
 & Clinical Knowledge & 264 \\
 & Security Studies & 244 \\
 & High School World History & 237 \\
 & Marketing & 234 \\
 & Conceptual Physics & 233 \\
 & High School Microeconomics & 233 \\
 & Human Aging & 223 \\
 & High School Statistics & 215 \\
 & High School US History & 204 \\
 & Sociology & 201 \\
 & High School Chemistry & 198 \\
 & High School Geography & 198 \\
 & High School Government and Politics & 193 \\
 & College Medicine & 172 \\
 & World Religions & 171 \\
 & Virology & 166 \\
 & High School European History & 164 \\
 & Logical Fallacies & 162 \\
 & High School Physics & 149 \\
 & Electrical Engineering & 145 \\
 & Astronomy & 145 \\
 & College Biology & 144 \\
 & Anatomy & 135 \\
 & Human Sexuality & 131 \\
 & Formal Logic & 126 \\
 & International Law & 121 \\
 & Econometrics & 114 \\
 & Machine Learning & 110 \\
 & Public Relations & 108 \\
 & Jurisprudence & 108 \\
 & Management & 103 \\
 & Medical Genetics & 100 \\
 & College Mathematics & 100 \\
 & Computer Security & 100 \\
 & Global Facts & 100 \\
 & Business Ethics & 100 \\
 & Abstract Algebra & 100 \\
 & College Computer Science & 100 \\
 & High School Computer Science & 100 \\
 & College Chemistry & 99 \\
 & US Foreign Policy & 99 \\
 & College Physics & 84 \\
\midrule

MedQA \cite{jin2020disease} &  & 1,273 \\
\midrule

\multirow{7}{*}{MATH \cite{hendrycksmath2021}}
 & Algebra & 1,187 \\
 & Intermediate Algebra & 903 \\
 & Prealgebra & 871 \\
 & Precalculus & 546 \\
 & Number Theory & 540 \\
 & Geometry & 479 \\
 & Counting \& Probability & 474 \\
\midrule

IFEval \cite{zhou2023instructionfollowingevaluationlargelanguage} &  & 541 \\
HellaSwag \cite{zellers-etal-2019-hellaswag} &  & 10,018 \\
GSM8K \cite{cobbe2021trainingverifierssolvemath} &  & 1,319 \\
\bottomrule
\end{tabular}
\end{table*}
\newpage

\begin{table*}
\centering
\scriptsize
\begin{tabular}{lcc}
\toprule
\textbf{Dataset} & \textbf{Task} & \textbf{Items} \\
\midrule

\multirow{2}{*}{GSM-Symbolic \cite{mirzadeh2025gsmsymbolic}}
 & P1 & 5,000 \\
 & P2 & 2,500 \\
\midrule

\multirow{6}{*}{Finance Fundamentals \cite{krumdick-etal-2024-bizbench}}
 & Finknow & 131 \\
 & TAT-QA (Extract) & 129 \\
 & CodeTAT-QA & 105 \\
 & ConvFinQA (Extract) & 94 \\
 & FinCode & 92 \\
 & CodeFinQA & 49 \\
\midrule

\multirow{2}{*}{CommonsenseQA \cite{talmor-etal-2019-commonsenseqa}}
 &  & 1,186 \\
 & Person & 35 \\
\midrule

\multirow{8}{*}{ChemBench \cite{Mirza2024AreLL}}
 & Toxicity and Safety & 674 \\
 & Organic Chemistry & 403 \\
 & Physical Chemistry & 165 \\
 & Analytical Chemistry & 152 \\
 & General Chemistry & 148 \\
 & Inorganic Chemistry & 92 \\
 & Materials Science & 83 \\
 & Technical Chemistry & 40 \\

\bottomrule
\end{tabular}
\end{table*}
\pagebreak
\pagebreak
\begin{table*}[ht]
\centering
\caption{List of models contained in WILD.}
\label{tab:wild_models}
\scriptsize
\begin{tabular}{lcr}
\toprule
\textbf{Model} & \textbf{Access?} \\
\midrule
Claude 3 Haiku & API \\
Claude 3 Sonnet & API \\
Claude 3.5 Haiku & API \\
Claude 3.5 Sonnet & API \\
GPT 4.1 & API \\
GPT 4.1 mini & API \\
GPT 4.1 nano & API \\
GPT-4o & API \\
GPT-4o mini & API \\
Nova Lite & API \\
Nova Pro & API \\
\midrule
meta-llama/Llama-3.1-70B & HuggingFace \\
meta-llama/Llama-3.1-8B & HuggingFace \\
meta-llama/Llama-3.2-1B & HuggingFace \\
meta-llama/Llama-3.2-3B & HuggingFace \\
01-ai/Yi-1.5-34B-Chat & HuggingFace \\
01-ai/Yi-1.5-6B-Chat & HuggingFace \\
01-ai/Yi-1.5-9B-Chat & HuggingFace \\
01-ai/Yi-34B-Chat & HuggingFace \\
01-ai/Yi-6B-Chat & HuggingFace \\
01-ai/Yi-Coder-9B-Chat & HuggingFace \\
CohereForAI/aya-expanse-32b & HuggingFace \\
CohereForAI/aya-expanse-8b & HuggingFace \\
CohereForAI/c4ai-command-r-plus-08-2024 & HuggingFace \\
CohereForAI/c4ai-command-r-v01 & HuggingFace \\
CohereForAI/c4ai-command-r7b-12-2024 & HuggingFace \\
Qwen/Qwen1.5-14B-Chat & HuggingFace \\
Qwen/Qwen1.5-32B-Chat & HuggingFace \\
Qwen/Qwen1.5-4B-Chat & HuggingFace \\
Qwen/Qwen1.5-7B-Chat & HuggingFace \\
Qwen/Qwen2-1.5B-Instruct & HuggingFace \\
Qwen/Qwen2-72B-Instruct & HuggingFace \\
Qwen/Qwen2-7B-Instruct & HuggingFace \\
Qwen/Qwen2-Math-72B-Instruct & HuggingFace \\
Qwen/Qwen2-Math-7B-Instruct & HuggingFace \\
Qwen/Qwen2.5-0.5B-Instruct & HuggingFace \\
Qwen/Qwen2.5-1.5B-Instruct & HuggingFace \\
Qwen/Qwen2.5-32B-Instruct & HuggingFace \\
Qwen/Qwen2.5-3B-Instruct & HuggingFace \\
Qwen/Qwen2.5-72B-Instruct & HuggingFace \\
Qwen/Qwen2.5-7B-Instruct & HuggingFace \\
Qwen/Qwen2.5-Coder-14B-Instruct & HuggingFace \\
Qwen/Qwen2.5-Coder-32B-Instruct & HuggingFace \\
allenai/OLMo-2-0325-32B-Instruct & HuggingFace \\
allenai/OLMo-2-1124-13B-Instruct & HuggingFace \\
allenai/OLMo-2-1124-7B-Instruct & HuggingFace \\
allenai/OLMo-7B-Instruct-hf & HuggingFace \\
deepseek-ai/deepseek-llm-7b-chat & HuggingFace \\
ibm-granite/granite-3.2-2b-instruct & HuggingFace \\
ibm-granite/granite-3.2-8b-instruct & HuggingFace \\
microsoft/phi-4 & HuggingFace \\
mistralai/Codestral-22B-v0.1 & HuggingFace \\
mistralai/Ministral-8B-Instruct-2410 & HuggingFace \\
mistralai/Mistral-7B-Instruct-v0.3 & HuggingFace \\
mistralai/Mistral-Small-Instruct-2409 & HuggingFace \\
mistralai/Mixtral-8x7B-Instruct-v0.1 & HuggingFace \\
nvidia/Nemotron-Mini-4B-Instruct & HuggingFace \\
swiss-ai/Apertus-70B-Instruct-2509 & HuggingFace \\
swiss-ai/Apertus-8B-Instruct-2509 & HuggingFace \\
tiiuae/Falcon3-10B-Instruct & HuggingFace \\
tiiuae/Falcon3-7B-Instruct & HuggingFace \\
utter-project/EuroLLM-22B-Instruct-2512 & HuggingFace \\
utter-project/EuroLLM-9B-Instruct & HuggingFace \\
zai-org/GLM-4-32B-0414 & HuggingFace \\
zai-org/GLM-4-9B-0414 & HuggingFace \\
\bottomrule
\end{tabular}

\end{table*}

\begin{figure*}[t]
  \centering
  \includegraphics[width=0.9\textwidth]{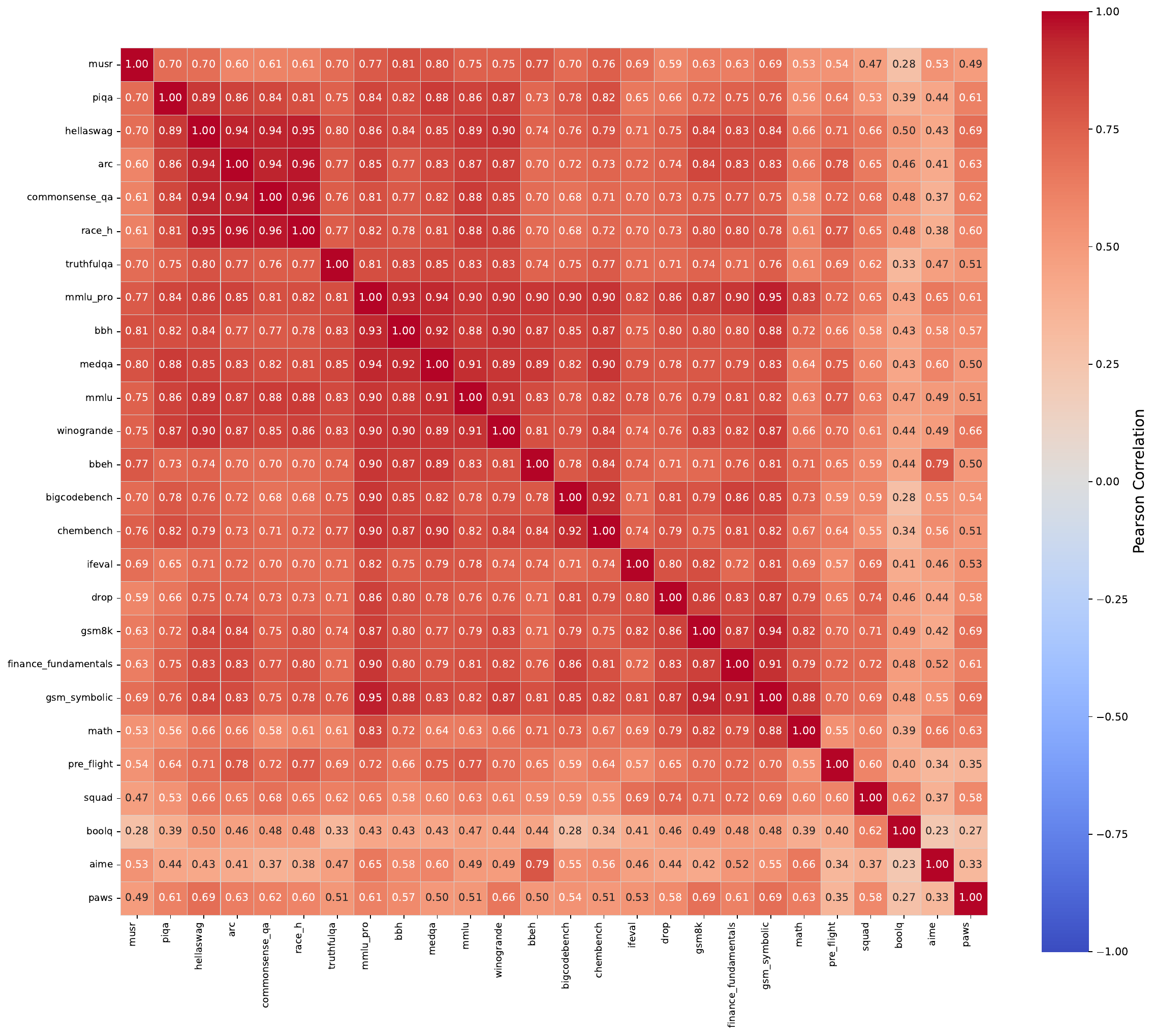}
  \caption{Pearson correlation between LLM performance across all base datasets. Datasets are ordered based on hierarchical clustering of their correlation strength.}
  \label{fig:base_task_corr}
\end{figure*}
\begin{figure*}[t]
  \centering
  \includegraphics[width=0.9\textwidth]{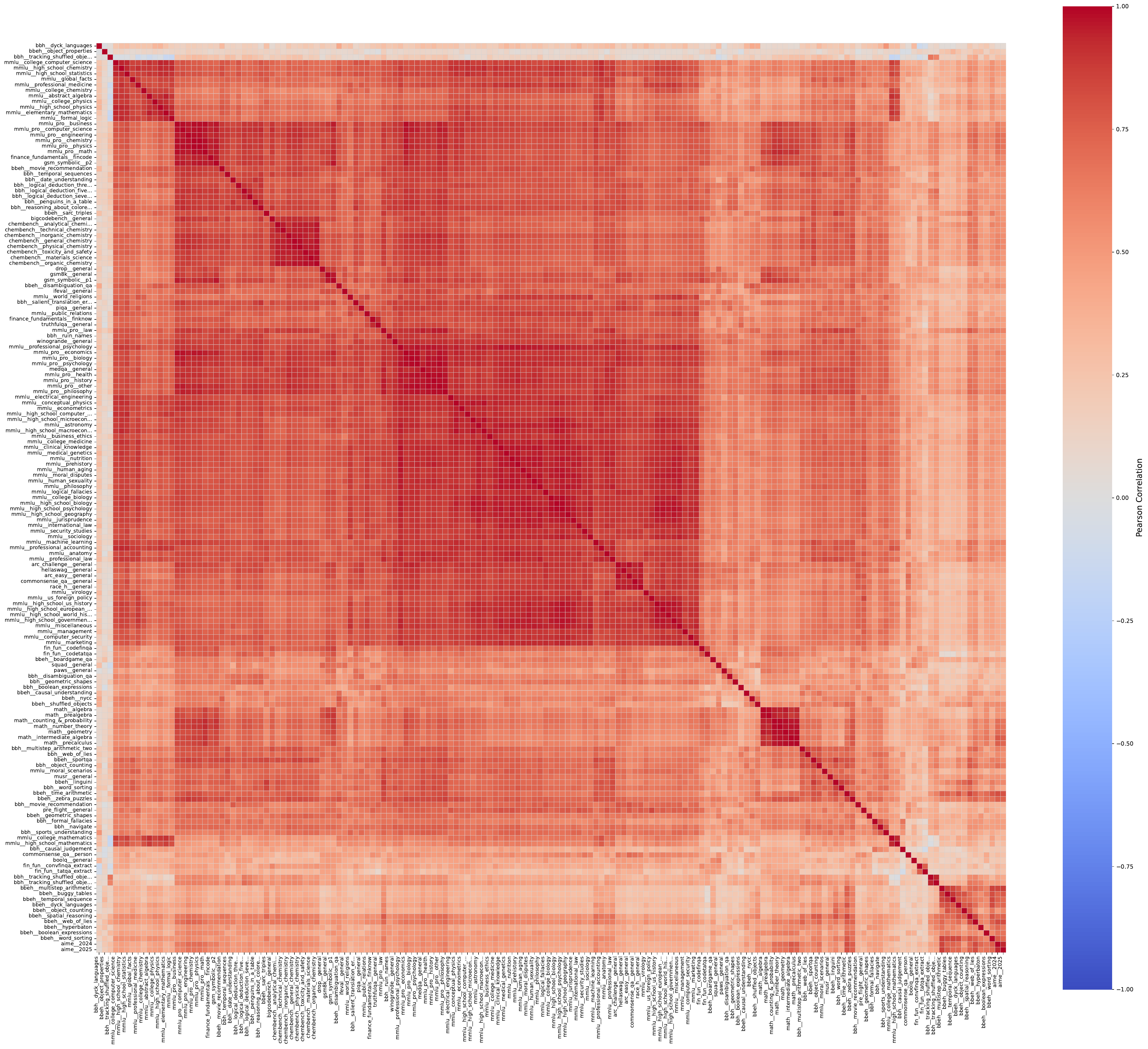}
  \caption{Pearson correlation between LLM performance across all sub-tasks. Tasks are ordered based on hierarchical clustering of their correlation strength.}
  \label{fig:sub_task_corr}
\end{figure*}

\clearpage

\end{document}